\pdfoutput=1
\documentclass[10pt,twocolumn]{article}
\usepackage[T1]{fontenc}
\usepackage[utf8]{inputenc}
\usepackage[letterpaper,margin=0.6in,columnsep=0.25in]{geometry}
\usepackage{newtxtext,newtxmath}
\usepackage{graphicx,booktabs,array,multirow,tabularx}
\usepackage{xcolor,amsmath}
\usepackage{microtype,placeins,float}
\usepackage{pgfplots}
\usetikzlibrary{patterns}
\pgfplotsset{compat=1.18}
\usepackage[backend=biber,style=numeric-comp,sorting=nyt,giveninits=true,maxbibnames=99]{biblatex}
\DeclareBibliographyAlias{preprint}{online}
\usepackage[hidelinks]{hyperref}
\providecommand{\Description}[2][]{}
\title{DiffLUT-Net: Differentiable Training of FPGA LUT Networks with Learnable Connectivity}
\author{%
\begin{tabular}{@{}ccc@{}}
Jiaqi Ye & Xinrui Gong & Jingcun Wang \\
{\small TU Darmstadt} & {\small TU Darmstadt} & {\small TU Darmstadt} \\
{\scriptsize\texttt{jiaqi.ye@stud.tu-darmstadt.de}} &
{\scriptsize\texttt{xinrui.gong@tu-darmstadt.de}} &
{\scriptsize\texttt{jingcun.wang@tu-darmstadt.de}} \\[1em]
Olga Kondrateva & Bing Li & Grace Li Zhang \\
{\small TU Darmstadt} & {\small TU Ilmenau} & {\small TU Darmstadt} \\
{\scriptsize\texttt{olga.kondrateva@kom.tu-darmstadt.de}} &
{\scriptsize\texttt{bing.li@tu-ilmenau.de}} &
{\scriptsize\texttt{grace.zhang@tu-darmstadt.de}}
\end{tabular}%
}
\date{}
\begin{document}
\maketitle
\begin{abstract}
Field-programmable gate arrays (FPGAs) enable efficient neural-network inference, but most deployment flows either accelerate multiply–accumulate operations or convert pretrained quantized models into lookup tables (LUTs). We present DiffLUT-Net, an FPGA-native network connected by six-input LUTs that are trained from scratch. We jointly learn the 64 truth-table entries of a LUT and the source to each of its six input ports using a differentiable LUT function relaxation and hardware source selection. After training, the truth tables and connections are discretized, unused logic can be pruned, and the network is exported directly as synthesizable Verilog. Across five benchmarks, DiffLUT-Net achieves favorable accuracy–resource trade-offs. A compact JSC CERNBox model reaches 72.5\% accuracy using 94 LUTs with 1.21 ns latency. On MNIST, a compact configuration reaches 98.0\% accuracy while using 41\% fewer LUTs than NeuraLUT-Assemble. On Fashion-MNIST, it matches DWN while using 52\% fewer LUTs; on CIFAR-10, it improves DWN accuracy by 2.08\% while using 16\% fewer LUTs. These results demonstrate the effectiveness of jointly learning LUT functions and sparse connectivity for compact FPGA-native inference. The code is available at \url{https://github.com/TUDa-HWAI/DiffLUT-Network}.
\end{abstract}


\section{Introduction}


Field-programmable gate arrays (FPGAs) are well suited to applications that require low and predictable inference latency. Their configurable logic, distributed memory, fine-grained parallelism, and customizable datapaths allow a trained model to be implemented as a deeply parallel hardware circuit. However, the efficiency of that circuit depends not only on model size or numerical precision, but also on how closely the trainable representation matches the primitives available in the FPGA fabric. In many existing design flows, model training and hardware realization remain separate: a network is first optimized using conventional multiply-accumulate (MAC) operations and is only later quantized, compiled, or mapped into FPGA resources. 
This separation creates a gap between the learned model and the logic that performs inference.


State-of-the-art research attempts to narrow this gap from two perspectives. 
First, truth-table-based methods convert trained quantized neurons or subnetworks into truth tables, which are implemented with LUTs. LogicNets, PolyLUT, NeuraLUT, and their extensions constrain the precision and fan-in of a neuron or compact subnetwork so that its input–output combinations can be enumerated after training and synthesized into FPGA LUTs~\cite{Umuroglu2020LogicNets, Andronic2023PolyLUT, Andronic2024NeuraLUT, Lou2024PolyLUTAdd, Weng2025AmigoLUT, Andronic2025NeuraLUTAssemble, Khataei2025TreeLUT, Hoang2026KANELE}. This strategy can absorb multiplication, accumulation, batch normalization, and non-linear activation into a truth table and thus LUTs. Its scalability is nevertheless limited by the exponential growth of a truth table with the number of input and output bits. 

Second, state-of-the-art research also defines the model directly as a network of Boolean gates or LUTs and trains these hardware-oriented functions from scratch. Differentiable logic gate networks use continuous relaxations to train small Boolean gates~\cite{Petersen2022DiffLogic}, while subsequent methods improve their parameterization, connectivity, architectural organization, and hardware realization~\cite{Ruttgers2025Light, Fojcik2025LILogicNet, Ma2026CLGN, Buehrer2026BitLogic}. Differentiable Weightless Neural Networks extend direct learning to multi-input LUTs and introduce learnable input mappings and output reductions~\cite{Bacellar2024DWN}. These studies show that logic functions can serve as trainable computational primitives rather than merely as post-training implementation targets.

The methodologies above are effective in many scenarios. However, they leave an important question unresolved: how can the Boolean functions and physical connectivity of an FPGA-native LUT network be optimized together while maintaining a direct path from training to hardware implementation? A physical LUT6 is defined by both its 64-entry truth table and the six signals connected to its inputs, and these choices are closely coupled. We therefore propose DiffLUT-Net, which jointly learns all truth-table entries and selects one source for each LUT input port. A differentiable relaxation enables function learning, while hard selection preserves FPGA-valid wiring. After training, the truth tables are binarized, the connections are fixed, unused logic can be pruned, and the resulting deterministic network is exported directly as synthesizable Verilog.

The main contributions of this work are:

\begin{itemize}
\item A LUT6-native differentiable model. We formulate the complete truth table of each six-input FPGA LUT as a closed-form multilinear relaxation. The formulation is exact for binary inputs and exposes all 64 truth-table entries as directly trainable parameters.
\item Joint optimization of LUT functions and hardware-valid sparse connectivity. Every LUT input port learns one source signal.  
After training, the learned scores become deterministic wiring and can expose unused upstream logic for synthesis-time removal. 
\item A complete training-to-hardware path. DiffLUT-Net integrates distribution-aware binary input encoding, one or more trainable LUT6 layers, grouped output accumulation, truth-table binarization, connection mapping, Verilog generation, and post-placement-and-routing FPGA evaluation.
\item A systematic experimental and architectural study. We evaluate DiffLUT-Net on five benchmarks against arithmetic accelerators, post-training truth-table methods, and directly trained logic or LUT networks. 
DiffLUT-Net demonstrates strong compact and high-accuracy operating points: at comparable accuracy, a compact MNIST configuration uses 41\% fewer LUTs than NeuraLUT\allowbreak-Assemble. Relative to DWN, DiffLUT-Net reduces LUT utilization by 52\% on Fashion-MNIST and by 16\% on CIFAR-10, while closely matching or improving classification accuracy.



\end{itemize}

The rest of this paper is organized as follows. Section 2 introduces the related work. Section 3 explains the proposed method. Sections 4 and 5 respectively present the
experimental results and conclusions.
\section{Related Work}
\label{sec:related_work}

Neural network implementations on FPGAs can be organized into three broad categories. In the first category, a conventional neural network is trained, and the FPGA implements and accelerates its multiply-accumulate (MAC) operations. In the second category, the inputs and outputs of a neuron or a group of neurons are trained and quantized to low bit widths so their functions can be treated as truth tables by enumerating all input-output combinations. Such truth tables are implemented directly with LUTs on FPGAs. 
In the third category, the trainable model is defined directly as a network of logic gates or LUTs, so hardware-native functions are optimized from scratch. 

\subsection{Direct Acceleration of MAC operations in Neural Networks}

The first category keeps the arithmetic structure of a conventional neural network. Fully connected and convolutional layers still compute weighted sums followed by activation functions. On the FPGA, these operations are implemented with multipliers, adders, accumulators, buffers, and pipelines. Designers adjust numerical precision, parallelism, pipeline depth, and operator reuse to balance resource use, latency, and throughput. Pruning \cite{10546870, 10247868}, quantization \cite{10137171}, early-exit \cite{10595861} and knowledge distillation \cite{10546606} can be used to reduce the number of multiply-accumulate operations. 

The hls4ml tool flow is a representative example. It translates trained neural networks into configurable high-level synthesis (HLS) designs  \cite{Fahim2021HLS4ML} and provides design parameters for numerical precision, arithmetic reuse, and pipelining, allowing the generated accelerator to be optimized for different latency, throughput, and resource constraints. The resulting hardware preserves the arithmetic structure of the original network: each neuron still computes a quantized weighted sum followed by an activation function, while HLS and downstream synthesis map the required multipliers, adders, accumulators, and buffers onto DSP blocks, FPGA LUTs, flip-flops, and on-chip memories.

Ngadiuba et al. use hls4ml to implement binary- and ternary-precision networks on FPGAs \cite{Ngadiuba2021HLS4MLCompression}. FINN implements binarized matrix-vector products with XNOR, population count, accumulation, and thresholding. Although these operators are inexpensive, each neuron still computes a quantized dot product \cite{Umuroglu2017FINN}. da4ml implements constant matrix-vector multiplications as multiplierless networks of shifts, additions, and subtractions that map to LUTs, while exactly preserving the quantized computation \cite{Sun2025DA4ML}.

These methods make neural-network arithmetic cheaper, but they do not change the basic form of the network. The FPGA still evaluates weighted sums and activations, while synthesis tools decide how these operations are mapped to physical LUTs. 
However, it may still consume substantial resources because it requires large adder trees and circuits for non-linear activation functions and batch normalization.  Besides, such methods must move and store many intermediate values.

\subsection{Implementation of Truth Tables for Quantized Neurons and Subnetworks }

This line of work trains and quantizes the inputs and outputs of a neuron or a group of neurons into low bit width and limits the fan-in of the neuron or the group of neurons. After training, the neuron or group of neurons can be converted into a truth table by enumerating all input-output combinations. Such a truth table can be directly synthesized into a circuit of physical FPGA LUTs.


LogicNets applies this concept to sparse and low-precision neurons. It puts the whole function of a neuron, including the weighted sum, normalization, and activation of each neuron, into a truth table and implements this truth table with LUTs, removing the explicit MAC operations and activation \cite{Umuroglu2020LogicNets}. PolyLUT uses the same conversion process but replaces the affine neuron with a multivariate polynomial \cite{Andronic2023PolyLUT}.  NeuraLUT considers the function of a small dense subnetwork as a truth table and implements it with LUTs. During training, the subnetwork uses full-precision internal layers, while its inputs and outputs are quantized. After training, the entire subnetwork is enumerated into a truth table and converted to LUTs \cite{Andronic2024NeuraLUT}. \cite{10473844} further converted pretrained quantized neurons into logic circuits by embedding fixed weights into full multipliers.

\begin{figure*}[!t]
  \centering
  \resizebox{\textwidth}{!}{\input{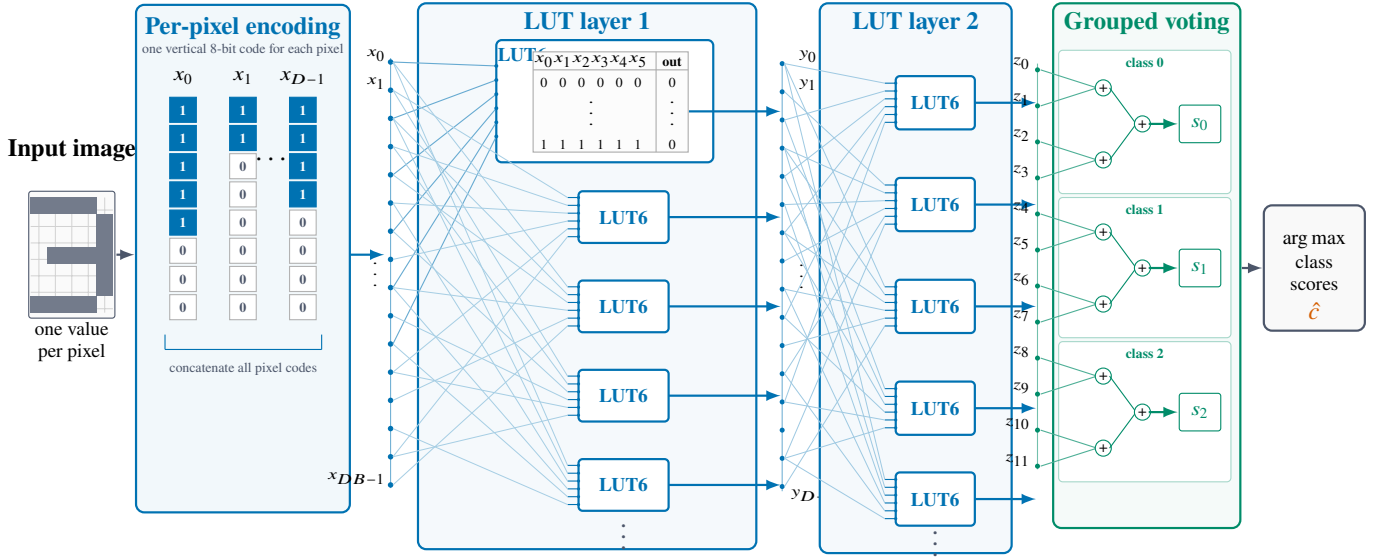}}
  \caption{Overview of the LUT networks.}
  \label{fig:overall}
  \Description{Overview of the proposed framework, including binary
  encoding, two LUT6 layers, class-score accumulation, and classification.}
\end{figure*}

The main limitation of such work is that the size of truth tables grows exponentially with the number and bit widths of fan-in. Subsequent methods therefore search for methods to increase computation capacity without enlarging one table.  PolyLUT-Add combines several small polynomial sub-neurons and adds their outputs to achieve high performance \cite{Lou2024PolyLUTAdd}.  NeuraLUT-Assemble increases neuron fan-in by assembling small NeuraLUT units into tree structures, with hardware-aware pruning used to group the inputs to these structures~\cite{Andronic2025NeuraLUTAssemble}. AmigoLUT improves model-level scalability by constructing ensembles of smaller LUT-based networks rather than continually enlarging individual neurons \cite{Weng2025AmigoLUT}. SparseLUT takes a complementary approach: under a fixed fan-in budget, it learns which inputs to keep instead of relying on a fixed random sparse pattern \cite{Lou2026SparseLUT}. KANEL{\' E} converts each one-dimensional spline on a Kolmogorov--Arnold Network edge into a LUT \cite{Hoang2026KANELE}.  TreeLUT is not a neural network: it maps trained gradient-boosted decision trees to pipelined FPGA logic \cite{Khataei2025TreeLUT}. 

Although the methods above are effective in many scenarios, the number of LUTs still increases exponentially with the increasing number of input bits. 
Consequently, these methods rely on low activation precision, sparse connectivity, restricted fan-in, decomposition, or hierarchical assembly.

\subsection{FPGA-Native LUT Networks and Hardware-Native Logic Networks}

The third category trains the hardware-native network from scratch. The model is built from Boolean logic gates or LUTs from scratch, rather than first training arithmetic neurons or subnetworks and converting them into LUTs afterward. Training chooses the function of each neuron and, in some methods, also its input connections.

Deep Differentiable Logic Gate Networks, denoted DiffLogic in our experiments, introduced a continuous relaxation for training a network consisting of two-input Boolean logic gates by representing their function as a distribution over the 16 possible gates 
\cite{Petersen2022DiffLogic}. Subsequent work improved this gate-level parameterization. WARP-LUTs replaces 16 variables with four Walsh-domain variables for a two-input gate \cite{Gerlach2025WARPLUTs}. Light Differentiable Logic Gate Networks develops a similar compact four-parameter representation and provides a detailed analysis of its effects on gradient propagation, discretization error, convergence, and trainability at depth \cite{Ruttgers2025Light}. 

Beyond gate parameterization, LILogicNet makes sparse gate connections trainable \cite{Fojcik2025LILogicNet}, while CLGN extends LUT-native learning to convolutional architectures and learns how LUT outputs are grouped for classification \cite{Ma2026CLGN}. BitLogic systematically develops a framework that compares encoders, connectivity rules, LUT fan-in, node parameterizations, and output heads under a shared protocol \cite{Buehrer2026BitLogic}. At the hardware-system level, FPGN extends differentiable functions to six-input LUTs, denoted as LUT6, and trains a LUT network with structured connectivity \cite{Liang2026FPGN}. 
Mommen et al. \cite{Mommen2026FullyTrainable} jointly optimizes LUT contents and LUT connections, whereas DWN \cite{Bacellar2024DWN} trains multi-input LUT contents using an extended finite-difference estimator and augments the network with learnable input mappings. 

 These LUT-level methods do not simultaneously establish the advantages of their LUT-training and connectivity choices and validate them through a complete training-to-Vivado workflow. Our work closes this gap through baseline comparisons and end-to-end hardware validation.

\section{Methodology of DiffLUT-Net}
\label{sec:methodology}


As illustrated in Figure~\ref{fig:overall}, DiffLUT-Net consists of three stages: thermometer encoding, LUT6-based computation, and grouped output accumulation. In the first stage, thermometer encoding converts real-valued features into binary signals, as described in Section~\ref{sec:Thermometer Encoding}. In the second stage, multiple LUT layers process and transform these binary signals, as described in Section~\ref{sec:Differentiable LUT Function} and Section~\ref{sec:Sparse Mapping}. In the third stage, GroupSum assigns LUT outputs in the last layer to class groups and predicts the class with the largest sum, as described in Section ~\ref{sec:grouped_output}. 

In the second stage, we first establish the LUT function to make it differentiable during training, as described in Section~\ref{sec:Differentiable LUT Function}. The connection between consecutive LUT layers is determined by learning a sparse mapping where each LUT in the later layer is connected to only 6 outputs from the former layer. This sparse mapping learning is described in Section ~\ref{sec:Sparse Mapping}.



\subsection{Differentiable LUT Function}
\label{sec:Differentiable LUT Function}

 A physical LUT operates on binary values, as shown in Table~\ref{tab:lut2_example}, where the truth table of a two-input LUT and its input representation are illustrated. The training goal of a LUT network is to determine the suitable storage values inside LUTs to maximize the accuracy of the LUT network on a given dataset.

\begin{table}[t]
\centering
\caption{Truth table of a two-input LUT and its input representation.}
\label{tab:lut2_example}
\small
\begin{tabular}{c c c c}
\toprule
\(x_0\) & \(x_1\) & stored value & input representation\\
\midrule
0 & 0 & \(\omega_{0}\) & \((1-x_0)(1-x_1)\) \\
0 & 1 & \(\omega_{1}\) & \((1-x_0)x_1\) \\
1 & 0 & \(\omega_{2}\) & \(x_0(1-x_1)\) \\
1 & 1 & \(\omega_{3}\) & \(x_0x_1\) \\
\bottomrule
\end{tabular}
\end{table}

To make a LUT trainable, we first express the function of the LUT as follows 
\begin{equation}
\begin{aligned}
y(x_0,x_1)
={}&
\omega_{0}(1-x_0)(1-x_1)
+\omega_{1}(1-x_0)x_1 \\
&+\omega_{2}x_0(1-x_1)
+\omega_{3}x_0x_1,
\end{aligned}
\label{eq:lut2_expanded}
\end{equation}
where $x_0$ and $x_1$ are the binary inputs and $\omega_{i}$ is the binary value stored for the $i$th address. For example, when $x_0=0$ and $x_1=0$, $y(0,0)=\omega_{0}$, which corresponds to the first part of the equation. 

$\omega_{i}$ should be restricted to binary values 0 or 1 since the output of a LUT is either 0 or 1. To achieve this goal, we used a sigmoid function and a trainable parameter   $\lambda_{j}$ to restrict the range of $\omega_{i}$ as follows 
\begin{equation}
\omega_{j}
=
\sigma(\lambda_{j}).
\label{eq:coefficient_parameterization}
\end{equation}
With this function, $\omega_{j}$ is restricted to $[0,1]$. For a 2-input LUT, the number of trainable parameters is 4. After training, each relaxed ${\omega}_{j}$ is converted into a binary value according to the following rules:
\begin{equation}
\widehat{\omega}_{j}
=
\begin{cases}
1, & \omega_{j}>0.5,\\
0, & \omega_{j}\leq 0.5,
\end{cases}
\label{eq:lut_hardening}
\end{equation}
where $\widehat{\omega}_{j}$, which is the binary value, is used in the inference.

\begin{figure}[t]
  \centering
\begin{tikzpicture}[
  x=1cm,y=1cm,
  font=\small,
  block/.style={draw=black!75,fill=black!2,rounded corners=2pt,line width=0.65pt},
  tableline/.style={draw=black!55,line width=0.45pt},
  signal/.style={-latex,draw=black!75,line width=0.6pt}
]

\draw[block] (0.85,0.25) rectangle (7.25,5.10);
\node[font=\small\bfseries] at (4.05,4.82) {$K$-input LUT};

\node[anchor=east] at (0.52,3.95) {$x_0$};
\node[anchor=east] at (0.52,3.20) {$x_1$};
\node at (0.34,2.40) {$\vdots$};
\node[anchor=east] at (0.52,1.42) {$x_{K-1}$};

\draw[signal] (0.58,3.95) -- (1.22,3.95);
\draw[signal] (0.58,3.20) -- (1.22,3.20);
\draw[signal] (0.58,1.42) -- (1.22,1.42);

\fill[blue!6] (1.35,4.00) rectangle (6.62,4.48);
\fill[blue!5] (1.35,2.03) rectangle (6.62,2.51);
\draw[tableline] (1.35,0.62) rectangle (6.62,4.48);
\draw[tableline] (1.35,4.00) -- (6.62,4.00);
\draw[tableline,line width=0.65pt] (4.98,0.62) -- (4.98,4.48);

\node at (1.78,4.24) {$x_0$};
\node at (2.60,4.24) {$x_1$};
\node at (3.43,4.24) {$\cdots$};
\node at (4.44,4.24) {$x_{K-1}$};
\node[font=\scriptsize\bfseries] at (5.80,4.24) {stored value};

\node at (1.78,3.67) {$0$};
\node at (2.60,3.67) {$0$};
\node at (3.43,3.67) {$\cdots$};
\node at (4.44,3.67) {$0$};
\node at (5.80,3.67) {$\omega_{0}$};

\node at (1.78,3.17) {$0$};
\node at (2.60,3.17) {$0$};
\node at (3.43,3.17) {$\cdots$};
\node at (4.44,3.17) {$1$};
\node at (5.80,3.17) {$\omega_{1}$};

\node at (1.78,2.73) {$\vdots$};
\node at (2.60,2.73) {$\vdots$};
\node at (3.43,2.73) {$\ddots$};
\node at (4.44,2.73) {$\vdots$};
\node at (5.80,2.73) {$\vdots$};

\node at (1.78,2.27) {$a_0$};
\node at (2.60,2.27) {$a_1$};
\node at (3.43,2.27) {$\cdots$};
\node at (4.44,2.27) {$a_{K-1}$};
\node at (5.80,2.27) {$\omega_{j}$};

\node at (1.78,1.79) {$\vdots$};
\node at (2.60,1.79) {$\vdots$};
\node at (3.43,1.79) {$\ddots$};
\node at (4.44,1.79) {$\vdots$};
\node at (5.80,1.79) {$\vdots$};

\node at (1.78,1.13) {$1$};
\node at (2.60,1.13) {$1$};
\node at (3.43,1.13) {$\cdots$};
\node at (4.44,1.13) {$1$};
\node at (5.80,1.13) {$\omega_{2^K-1}$};

\draw[signal] (6.62,2.27) -- (8.02,2.27);
\node[anchor=west] at (8.08,2.27) {$y$};

\end{tikzpicture}
  \caption{Truth table of a \(K\)-input LUT. 
  The highlighted row \(a_0,\ldots,a_{K-1}\) denotes an arbitrary input entry and stored value \(\omega_{j}\).} 
  \label{fig:lut_truth_table}
  \Description{A K-input LUT contains a truth table whose columns are the input positions. An arbitrary address row a zero through a K minus one stores coefficient omega j a and produces output y j when selected.}
\end{figure}
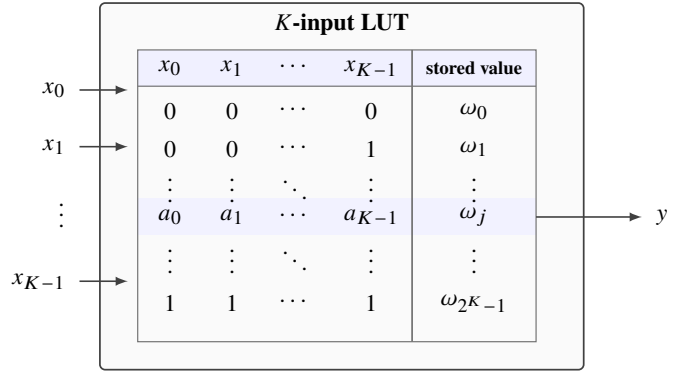

 The function of a 2-input LUT can be extended to a $K$-input LUT, e.g., $K=6$ in many FPGAs. The truth table of a $K$-input LUT is illustrated in Figure~\ref{fig:lut_truth_table}. In such a truth table, 
we used \((a_0,\ldots,a_{K-1})\),  where \(a_i\in\{0,1\}\) is the value at the $i$th position, to represent one input entry. When 
$a_i=0$ and $a_i=1$, the $i$th position can be expressed as $1-x_i$ and $x_i$, respectively. Accordingly, we express the input representation of the entry \((a_0,\ldots,a_{K-1})\) as follows 
\begin{equation}
\prod_{i=0}^{K-1}x_i^{a_i}(1-x_i)^{1-a_i}.
\label{eq:lut_address_selector}
\end{equation}
With the representation above, for example, when $a_i=0$ where $i=0,...,K-1$, the representation becomes $\prod_{i=0}^{K-1}(1-x_i)$. 
 
 According to the input representation above, we can establish the function of a $K$-input LUT as follows 
\begin{equation}
\begin{aligned}
y(x_0,\ldots,x_{K-1})
=
\sum_{u=0}^{2^K-1}
\omega_{j}
\prod_{i=0}^{K-1}
x_i^{a_i}(1-x_i)^{1-a_i},
\end{aligned}
\label{eq:differentiable_lut}
\end{equation}
where $\omega_{j}=\sigma(\lambda_{j})$, to make the stored values in the LUT restricted to \([0,1]\), similar to that in 2-input LUT function and $u$ indicates the address of input entries.  A LUT6 has 64 trainable parameters. The model learns these 64 parameters, i.e., \(\lambda_0,\ldots,\lambda_{63}\), directly.





\subsection{Sparse Connection between Consecutive LUT Layers}
\label{sec:Sparse Mapping}

In this section, we develop a strategy to make the connection between consecutive layers trainable, indicating that each LUT input port of the current layer can select its source from the previous layer during training. The same method also determines the connections between the encoded inputs and the first LUT layer.

\begin{figure*}[t]
  \centering
\begin{tikzpicture}[
  x=1cm,y=1cm,
  font=\scriptsize,
  layerbox/.style={draw=black!75,rounded corners=1.5pt,line width=0.55pt},
  currentbox/.style={draw=black!75,dashed,rounded corners=1.5pt,line width=0.55pt},
  lut/.style={draw=black!80,fill=black!3,rounded corners=1pt,
              minimum width=0.82cm,minimum height=0.46cm,inner sep=1pt},
  lutlarge/.style={draw=black!80,fill=black!3,rounded corners=1pt,
                   minimum width=0.88cm,minimum height=1.18cm,inner sep=1pt},
  inactive/.style={draw=black!20,text=black!30,fill=black!2},
  pin/.style={circle,fill=black!80,inner sep=0.75pt},
  ipin/.style={circle,fill=black!22,inner sep=0.75pt},
  learn/.style={draw=blue!62,opacity=0.18,line width=0.30pt},
  fixed/.style={draw=blue!72!black,line width=0.68pt},
  guide/.style={draw=black!25,line width=0.35pt}
]

\node[anchor=west,font=\small\bfseries] at (0.00,5.05) {(a) Learnable connections};

\draw[layerbox] (0.05,0.45) rectangle (1.52,4.65);
\node[font=\scriptsize\bfseries] at (0.79,4.42) {Previous layer};

\node[lut] at (0.75,3.85) {LUT6};
\node[lut] at (0.75,3.15) {LUT6};
\node at (0.75,2.50) {$\vdots$};
\node[lut] at (0.75,1.85) {LUT6};
\node[lut] at (0.75,1.15) {LUT6};

\foreach \yy in {3.85,3.15,1.85,1.15}{
  \node[pin] at (1.52,\yy) {};
  \draw[black!65,line width=0.35pt] (1.52,\yy) -- (1.98,\yy);
}
\node[anchor=west,fill=white,inner sep=0.5pt] at (1.62,3.85) {$y_0$};
\node[anchor=west,fill=white,inner sep=0.5pt] at (1.62,3.15) {$y_1$};
\node[anchor=west,fill=white,inner sep=0.5pt] at (1.60,1.85) {$y_{D_{\mathrm{in}}-2}$};
\node[anchor=west,fill=white,inner sep=0.5pt] at (1.60,1.15) {$y_{D_{\mathrm{in}}-1}$};

\draw[currentbox] (3.65,0.45) rectangle (5.10,4.65);
\node[font=\scriptsize\bfseries] at (4.40,4.42) {Current layer};

\node[lutlarge] at (4.42,3.55) {$\mathrm{LUT6}_0$};
\node[lutlarge] at (4.42,2.00) {$\mathrm{LUT6}_1$};
\node at (4.42,0.91) {$\vdots$};

\foreach \yy in {4.05,3.85,3.65,3.45,3.25,3.05}
  \node[pin] at (3.96,\yy) {};
\foreach \yy in {2.50,2.30,2.10,1.90,1.70,1.50}
  \node[pin] at (3.96,\yy) {};
\node[pin] at (4.88,3.55) {};
\node[pin] at (4.88,2.00) {};
\draw (4.88,3.55) -- (5.10,3.55);
\draw (4.88,2.00) -- (5.10,2.00);

\foreach \sy in {3.85,3.15,1.85,1.15}{
  \foreach \ty in {4.05,3.85,3.65,3.45,3.25,3.05,2.50,2.30,2.10,1.90,1.70,1.50}{
    \draw[learn] (1.98,\sy) -- (3.96,\ty);
  }
}
\node[text=blue!62!black] at (2.68,0.72) {trainable scores};

\node[anchor=west,font=\small\bfseries] at (5.42,5.05) {(b) Score matrix $A=[a_{i,j}]$};
\node[align=center,text=black!75] at (8.88,4.65)
  {candidate outputs from the preceding layer};

\fill[blue!6,rounded corners=1pt] (5.98,2.10) rectangle (11.82,4.10);
\draw[blue!58,line width=0.65pt,rounded corners=1pt] (5.98,2.10) rectangle (11.82,4.10);

\draw[black!65,line width=0.5pt] (5.98,0.52) rectangle (11.82,4.40);
\draw[guide] (6.98,0.52) -- (6.98,4.40);
\draw[guide] (5.98,4.10) -- (11.82,4.10);
\node at (7.62,4.25) {$y_0$};
\node at (8.72,4.25) {$y_1$};
\node at (9.82,4.25) {$\cdots$};
\node at (11.08,4.25) {$y_{D_{\mathrm{in}}-1}$};

\node at (6.53,3.92) {$x_0$};
\node at (6.53,3.58) {$x_1$};
\node at (6.53,3.24) {$x_2$};
\node at (6.53,2.90) {$x_3$};
\node at (6.53,2.56) {$x_4$};
\node at (6.53,2.22) {$x_5$};
\node at (6.53,1.62) {$\vdots$};
\node[font=\tiny] at (6.53,0.86) {$x_{6D_{\mathrm{out}}-1}$};

\node at (7.62,3.92) {$a_{0,0}$};
\node at (8.72,3.92) {$a_{0,1}$};
\node at (9.82,3.92) {$\cdots$};
\node at (11.08,3.92) {$a_{0,D_{\mathrm{in}}-1}$};

\node at (7.62,3.58) {$a_{1,0}$};
\node at (8.72,3.58) {$a_{1,1}$};
\node at (9.82,3.58) {$\cdots$};
\node at (11.08,3.58) {$a_{1,D_{\mathrm{in}}-1}$};

\node at (7.62,3.24) {$a_{2,0}$};
\node at (8.72,3.24) {$a_{2,1}$};
\node at (9.82,3.24) {$\cdots$};
\node at (11.08,3.24) {$a_{2,D_{\mathrm{in}}-1}$};

\node at (7.62,2.90) {$a_{3,0}$};
\node at (8.72,2.90) {$a_{3,1}$};
\node at (9.82,2.90) {$\cdots$};
\node at (11.08,2.90) {$a_{3,D_{\mathrm{in}}-1}$};

\node at (7.62,2.56) {$a_{4,0}$};
\node at (8.72,2.56) {$a_{4,1}$};
\node at (9.82,2.56) {$\cdots$};
\node at (11.08,2.56) {$a_{4,D_{\mathrm{in}}-1}$};

\node at (7.62,2.22) {$a_{5,0}$};
\node at (8.72,2.22) {$a_{5,1}$};
\node at (9.82,2.22) {$\cdots$};
\node at (11.08,2.22) {$a_{5,D_{\mathrm{in}}-1}$};

\node at (7.62,1.62) {$\vdots$};
\node at (8.72,1.62) {$\vdots$};
\node at (9.82,1.62) {$\ddots$};
\node at (11.08,1.62) {$\vdots$};

\node[font=\tiny] at (7.58,0.86) {$a_{6D_{\mathrm{out}}-1,0}$};
\node at (9.38,0.86) {$\cdots$};
\node[font=\tiny] at (11.03,0.86) {$a_{6D_{\mathrm{out}}-1,D_{\mathrm{in}}-1}$};

\draw[blue!72!black,line width=0.7pt]
  (5.88,4.06) -- (5.78,4.06) -- (5.78,2.16) -- (5.88,2.16);
\node[rotate=90,text=blue!72!black,font=\scriptsize\bfseries]
  at (5.48,3.11) {six ports of $\mathrm{LUT6}_0$};

\node[anchor=west,font=\small\bfseries] at (11.92,5.05) {(c) Fixed connections};

\draw[layerbox] (12.00,0.45) rectangle (13.55,4.65);
\node[font=\scriptsize\bfseries] at (12.78,4.42) {Previous layer};

\node[lut] at (12.68,3.92) {LUT6};
\node[lut,inactive] at (12.68,3.35) {LUT6};
\node[lut] at (12.68,2.78) {LUT6};
\node at (12.68,1.98) {$\vdots$};
\node[lut,inactive] at (12.68,0.86) {LUT6};

\node[pin] at (13.55,3.92) {};
\node[ipin] at (13.55,3.35) {};
\node[pin] at (13.55,2.78) {};
\node[ipin] at (13.55,0.86) {};
\foreach \yy in {2.35,2.05,1.75,1.45}{
  \draw[black!45,line width=0.35pt] (13.18,\yy) -- (13.55,\yy);
  \node[pin] at (13.55,\yy) {};
}

\node[anchor=west,fill=white,inner sep=0.5pt] at (13.65,3.92) {$y_0$};
\node[anchor=west,fill=white,inner sep=0.5pt,text=black!30] at (13.65,3.35) {$y_1$};
\node[anchor=west,fill=white,inner sep=0.5pt] at (13.65,2.78) {$y_2$};
\node[anchor=west,fill=white,inner sep=0.5pt,text=black!30] at (13.62,0.86) {$y_{D_{\mathrm{in}}-1}$};

\draw[currentbox] (15.72,0.45) rectangle (17.20,4.65);
\node[font=\scriptsize\bfseries] at (16.46,4.42) {Current layer};

\node[lutlarge,minimum height=1.50cm] at (16.48,3.15) {$\mathrm{LUT6}_0$};
\foreach \yy in {3.70,3.48,3.26,3.04,2.82,2.60}
  \node[pin] at (16.00,\yy) {};
\node[pin] at (16.96,3.15) {};
\draw (16.96,3.15) -- (17.10,3.15);

\draw[fixed] (13.98,3.92) -- (16.00,3.70);
\draw[fixed] (13.98,2.78) -- (16.00,3.48);
\draw[fixed] (13.55,2.35) -- (16.00,3.26);
\draw[fixed] (13.55,2.05) -- (16.00,3.04);
\draw[fixed] (13.55,1.75) -- (16.00,2.82);
\draw[fixed] (13.55,1.45) -- (16.00,2.60);

\draw[black!65,line width=0.35pt] (13.55,3.92) -- (13.98,3.92);
\draw[black!65,line width=0.35pt] (13.55,2.78) -- (13.98,2.78);

\node at (16.48,1.42) {$\vdots$};
\foreach \yy in {1.72,1.42,1.12}
  \node[pin] at (16.00,\yy) {};
\draw[fixed,opacity=0.46] (13.55,2.35) -- (16.00,1.72);
\draw[fixed,opacity=0.46] (13.55,2.05) -- (16.00,1.42);
\draw[fixed,opacity=0.46] (13.55,1.75) -- (16.00,1.12);

\draw[black!15] (5.30,0.28) -- (5.30,5.12);
\draw[black!15] (11.88,0.28) -- (11.88,5.12);

\end{tikzpicture}
  \caption{Learnable mapping between two LUT layers. (a) During training, each input port of a LUT in the current layer considers all the output ports from the previous layer. (b) The connection matrix $A$.  The highlighted first six rows indicate the six input ports of the first LUT6 in the current layer. (c) After training, one source is retained for every input port in the current layer. LUTs in the previous layer that are not selected by any input port in the current layer become inactive and are pruned. }
  \label{fig:learnable_mapping_overview}
  \Description{Three panels show learnable connections, their score-matrix representation, and the fixed sparse wiring obtained after row-wise selection.}
\end{figure*}
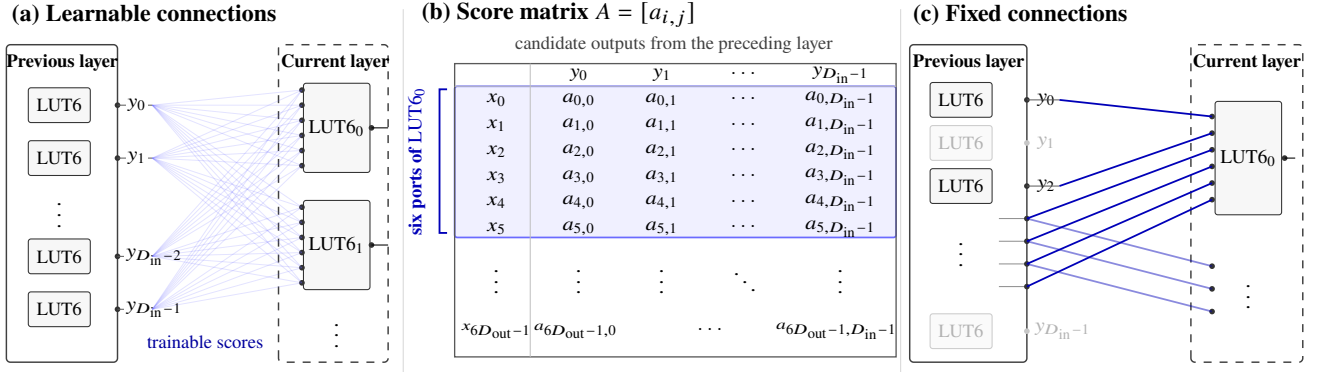

Figure~\ref{fig:learnable_mapping_overview}(a) illustrates the connection between two consecutive LUT layers. The previous layer provides \(D_{\mathrm{in}}\) output binary signals \(y_0,\ldots,y_{D_{\mathrm{in}}-1}\).  \(D_{\mathrm{out}}\) denotes the number of LUTs in the current layer. Since every LUT in the current layer has six input ports, the current layer contains \(6D_{\mathrm{out}}\) input ports, denoted by \(x_0,\ldots,x_{6D_{\mathrm{out}}-1}\).

 To determine which output port in the previous layer connects to which input port in the current layer, we construct the connection matrix $A$, as illustrated in Figure~\ref{fig:learnable_mapping_overview}(b). In this matrix, each row indicates the connection from all output ports of the previous layer to the input port $x_i$ of the current layer. Each column indicates all the connections from the output port $y_j$ of the previous layer to the current layer. The entry $a_{i,j}$ is the trainable parameter connecting $y_j$ to $x_i$. The first six rows, $x_0,\ldots,x_5$, correspond to the six input ports of the first LUT6 in the current layer; the next six rows correspond to the second node, etc.
 



 To train the connection matrix $A$, in the forward propagation, for the input port $x_i$ in the current layer, 
  the output port among $y_0,..., y_{D_{in}-1}$ in the $i$th row that has the largest value is used as the input port of $x_i$ in the current layer, expressed as follows:
\begin{equation}
m_i
=
\arg\max_{0\leq j<D_{\mathrm{in}}} a_{i,j},
\qquad
x_i
=
y_{m_i},
\label{eq:hard_mapping}
\end{equation}
where $m_i$ is the column that has the largest value in the $i$th row. 

 In backpropagation, the 
 \(\arg\max\) is not differentiable. To address this issue, we exploit the heuristic parameter update during training in DWN~\cite{Bacellar2024DWN}. Specifically, for one training sample, let
\(
g_i=\partial\mathcal{L}_{\mathrm{CE}}/\partial x_i
\)
be the gradient arriving at the port \(x_i\). With learning rate \(\eta\), every score in the $i$th row is updated as follows
\begin{equation}
\begin{aligned}
\frac{\partial\mathcal{L}_{\mathrm{CE}}}
     {\partial a_{i,j}}
&=
(2y_j-1)g_i, \\
a_{i,j}
&\leftarrow
a_{i,j}-\eta(2y_j-1)g_i,
\end{aligned}
\label{eq:mapping_weight_gradient}
\end{equation}
where $y_j$ is the $j$th output from the previous layer. 
The sign of \(g_i\) indicates whether a larger or smaller value should be optimized for \(x_i\) to minimize the loss function. Equation~(\ref{eq:mapping_weight_gradient}) therefore increases the connection scores of candidates that follow this direction and decreases the remaining values. In each training iteration, all candidates are updated, although only the candidate with the largest score is used in the forward propagation. 

After training, the largest entry in each row of the connection
matrix determines the final source for the corresponding input port.
For input port $x_i$, the selected column index $m_i$ specifies
the connection from output $y_{m_i}$ of the previous layer,
as shown in Figure~\ref{fig:learnable_mapping_overview}(c).
A LUT in the previous layer whose output is not selected by any
input port in the current layer becomes inactive and can be pruned.

\subsection{Training A LUT Network}
\label{sec:Training A LUT Network}


With the differentiable LUT function in Section 3.1 and the learnable connection in Section 3.2, in this section, we will introduce how we train a LUT network from scratch. For the initialization of \(\lambda\), the \(\lambda\) of each LUT independently takes either \(+\lambda_{\mathrm{init}}\) or \(-\lambda_{\mathrm{init}}\) with equal probability. In the experiments, \(+\lambda_{\mathrm{init}}\) and \(-\lambda_{\mathrm{init}}\) are set to 10 and -10, respectively. This symmetric signed initialization enables more informative gradients to propagate to early layers and thus facilitates faster convergence in practice \cite{Ruttgers2025Light}. 

To avoid overfitting during training, in the cost function, which typically includes a cross-entropy function, we add another penalty to minimize the distance between \(\lambda\) and 0. This penalty is similar to \(L_2\) regularization to push the value of \(\lambda\) to be 0 as much as possible.

After training, each relaxed parameter ${\omega}_{j}=\sigma(\lambda_{j})$ is converted into a binary value:
\begin{equation}
\widehat{\omega}_{j}
=
\begin{cases}
1, & \omega_{j}>0.5,\\
0, & \omega_{j}\leq 0.5.
\end{cases}
\label{eq:lut_hardening}
\end{equation} 

Once the stored values of each LUT and all connections between LUTs are determined, the complete  LUT network is determined. The LUT network can then be translated directly into RTL, exported as Verilog, and synthesized and implemented with Vivado.

\subsection{Thermometer Encoding for Input Data}
\label{sec:Thermometer Encoding}

 This section determines the encoding of input data. The inputs to the first LUT layer should be binary values. However, the input data, such as a pixel in an input image, are usually real values, e.g., 0.73. 
 Intuitively, quantizing inputs into 8 bits can convert inputs into binary values. However, this quantization cannot reflect the actual distribution of input features among all training data.  
 To address this issue, we adopt the Distributive Thermometer encoding~\cite{Bacellar2022DistributiveThermometer}. Specifically, for a feature \(x_i\), we compute \(B\) ordered thresholds from the training data as follows
 
\begin{equation}
\theta_{i,j}
=
Q_i\!\left(\frac{j}{B+1}\right),
\qquad
j\in\{1,\ldots,B\},
\label{eq:thermometer_thresholds}
\end{equation}
where $\theta_{i,j}$ is the $j$th threshold for the feature or input pixel $x_i$. \(Q_i\) is the distribution of $x_i$, which is obtained by evaluating the value of $x_i$ among all training data. \(Q_i(p)\) is the empirical \(p\)-quantile of the probability distribution of feature \(i\).  

  The thresholds are estimated once from the training data and then fixed. Inference therefore requires only comparisons before entering the LUT network. 
 For example, when $B=8$, the eight quantile thresholds divide the $i$th pixel value distribution among training data into nine regions, which consist of approximately equal numbers of training samples. This distribution-aware partitioning generates more thresholds where feature values occur more frequently.

 

By comparing the pixel values with each threshold, we could generate one binary value as follows 
\begin{equation}
x_{i,j}
=
\begin{cases}
1, & x_i\geq\theta_{i,j},\\
0, & x_i<\theta_{i,j},
\end{cases}
\label{eq:thermometer_bit}
\end{equation}
where $x_{i,j}$ is the binary value after being compared with the $j$th threshold.

\begin{figure}[t]
  \centering
  \definecolor{encblue}{RGB}{0,114,178}
\begin{tikzpicture}[x=1cm,y=1cm,font=\footnotesize]
  \node[font=\scriptsize\bfseries] at (0.53,2.30) {input image};
  \draw[fill=gray!4,draw=gray!65] (0.08,0.72) rectangle (0.98,1.98);
  \draw[step=0.18cm,gray!35,very thin] (0.08,0.72) grid (0.98,1.98);
  \foreach \col/\row in {
    0/6,1/6,2/6,3/6,
    4/5,4/4,
    1/3,2/3,3/3,4/3,
    4/2,4/1,
    0/0,1/0,2/0,3/0}
  {
    \fill[black!68]
      ({0.08+0.18*\col},{0.72+0.18*\row})
      rectangle ++(0.18,0.18);
  }

  \fill[encblue!72]
    ({0.08+0.18*4},{0.72+0.18*4})
    rectangle ++(0.18,0.18);
  \draw[encblue,very thick]
    ({0.08+0.18*4},{0.72+0.18*4})
    rectangle ++(0.18,0.18);
  \draw[encblue!70,thin] (0.98,1.44) -- (1.22,1.20);
  \draw[encblue!70,thin] (0.98,1.62) -- (1.22,1.90);
  \draw[fill=encblue!12,draw=encblue,thick]
    (1.22,1.20) rectangle (1.92,1.90);
  \node[font=\normalsize\bfseries,encblue] at (1.57,1.55) {0.73};
  \node[font=\scriptsize,align=center] at (1.57,0.96) {pixel \(i\)\\\(x_i=0.73\)};

  \draw[-latex,gray!75,thick] (1.96,1.55) -- (2.48,1.55);
  \begin{scope}[xshift=0.5cm]
  \draw[-latex,gray!75] (2.22,0.65) -- (5.72,0.65);
  \draw[-latex,gray!75] (2.22,0.65) -- (2.22,2.35);
  \node[font=\tiny,rotate=90] at (2.06,1.48) {frequency};
  \node[font=\tiny] at (3.75,0.32) {pixel threshold};

  \foreach \tx in {2.48,2.67,2.86,3.06,3.29,3.83,4.38,5.05}
    \draw[gray!58,dashed,thin] (\tx,0.66) -- (\tx,2.20);

  \draw[black!85,thick]
    plot[smooth] coordinates {
      (2.22,0.66) (2.36,0.72) (2.50,0.92) (2.70,1.46)
      (2.93,2.08) (3.18,2.35) (3.42,2.13) (3.67,1.75)
      (3.98,1.42) (4.35,1.16) (4.78,0.95) (5.25,0.78)
      (5.62,0.68)
    };

  \node[font=\tiny,anchor=north] at (2.48,0.61) {\(\theta_{i,1}\)};
  \node[font=\tiny,anchor=north] at (3.76,0.61) {\(\cdots\)};
  \node[font=\tiny,anchor=north] at (5.05,0.61) {\(\theta_{i,8}\)};
  \draw[encblue,very thick] (3.56,0.65) -- (3.56,2.28);
  \node[font=\scriptsize\bfseries,encblue,fill=white,inner sep=1.2pt]
    at (3.86,2.39) {\(x_i=0.73\)};
  \end{scope}

  \draw[-latex,gray!75,thick] (5.67,1.50) -- (5.94,1.50);
  \node[font=\scriptsize\bfseries] at (7.12,2.06) {thermometer code \(x_i\)};
  \foreach \bit [count=\idx from 0] in {1,1,1,1,1,0,0,0}
  {
    \pgfmathsetmacro{\bx}{6.00+0.28*\idx}
    \ifnum\bit=1
      \draw[fill=encblue,draw=encblue]
        (\bx,1.25) rectangle ++(0.27,0.48);
      \node[white,font=\scriptsize\bfseries] at ({\bx+0.135},1.49) {\bit};
    \else
      \draw[fill=gray!10,draw=gray!65]
        (\bx,1.25) rectangle ++(0.27,0.48);
      \node[black!80,font=\scriptsize\bfseries] at ({\bx+0.135},1.49) {\bit};
    \fi
  }
\end{tikzpicture}
  \caption{The concept of thermometer encoding where each pixel uses 8 bits to represent its value. The sample value \(x_i=0.73\) is larger than five thresholds, so its encoding is \(11111000\).  }
  \label{fig:thermometer_encoding}
  \Description{A pixelated digit three contains one highlighted pixel with value 0.73. The value is compared with eight quantile thresholds estimated from the same pixel position over all training images, producing the eight-bit thermometer code 11111000.}
\end{figure}
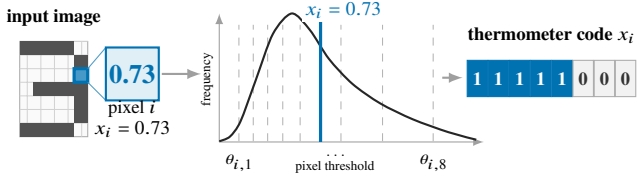

Figure~\ref{fig:thermometer_encoding} illustrates the concept of this input encoding, where 8-bit encoding is used. 
In this figure, the highlighted pixel in the sample image has a real value \(x_i=0.73\). This value is larger than the first five thresholds but smaller than the remaining three. Therefore, it is encoded as \(11111000\). The other input pixels can be encoded in a similar way. After all the input pixels are encoded into binary values, these encoded bits are concatenated and used as the candidate inputs for the first LUT layer.  

\subsection{Grouped Output Accumulation}
\label{sec:grouped_output}

GroupSum in the proposed method is responsible for converting the outputs of the final LUT layer into the scores of classes. To make the training robust, we use the accumulation of several LUT outputs as the score for one class.

Assume \(y_0,\ldots,y_{n_{\mathrm{out}}-1}\) denote the outputs of  the last layer and $n_{cls}$ is the total number of  classes.  Accordingly, the number of LUT outputs to obtain the score of one class is $p={n_{out}}/n_{cls}$.
Let \(\mathcal{G}_c\) denote the set of output indices assigned to class \(c\).  
Accordingly, the accumulated score for this class is the sum of the LUT outputs:
\begin{equation}
s_c
=
\sum_{j\in\mathcal{G}_c}
y_j.
\label{eq:groupsum}
\end{equation}
For example, if the final layer has 12 outputs and the task has three classes, four outputs accumulate for each class. Adding these four values generates the corresponding class score.



 During training, the accumulated class scores are divided by a temperature \(\tau\) to generate the logits used by the cross-entropy loss as follows. The division by \(\tau\) can narrow the gap in outputs of different classes to avoid the case that small outputs lead to negligible probabilities.  \
\begin{equation}
\ell_c
=
\frac{s_c}{\tau}.
\label{eq:groupsum_logits}
\end{equation} 

 In inference, the classification result is the group with the largest accumulated vote:
\begin{equation}
\widehat{c}
=
\arg\max_{0\leq c<n_{\mathrm{cls}}} s_c ,
\label{eq:groupsum_prediction}
\end{equation}
where $\widehat{c}$ is the classification result.

\section{Experimental Results}

\newcommand{\pslutCernFourKTwoK}{1257 and 2000}
\newcommand{\pslutOpenmlTwoKFiveHundred}{578 and 500}
\newcommand{\pslutMnistTwoKOneK}{695 and 999}
\newcommand{\pslutFashionTwoKOneK}{831 and 1000}

We evaluate whether jointly learning LUT6 functions and sparse connectivity produces accurate, resource-efficient circuits after the complete training-to-Verilog flow. The five benchmarks cover compact tabular models and larger image classifiers, with post-placement-and-routing hardware metrics reported where available. Because FPGA deployment is a multi-objective problem, we report compact configurations (denoted as low-accuracy in the result tables) to evaluate hardware efficiency at a matched accuracy and high-accuracy configurations (denoted as high-accuracy in the result tables) to characterize the accuracy obtained with a higher resource budget.


\subsection{Experimental Setup}

JSC OpenML~\cite{Duarte2018FastInference,OpenML2020HLS4MLJets} and JSC CERNBox~\cite{Duarte2018FastInference,CERN2025LHCJets} are five-class jet-classification tasks with 16 input features; MNIST~\cite{Deng2012MNIST}, Fashion-MNIST~\cite{Xiao2017FashionMNIST}, and CIFAR-10~\cite{Krizhevsky2009CIFAR10} are image classification datasets. JSC and MNIST designs target the Xilinx Virtex UltraScale+ \texttt{xcvu9p-\allowbreak flgb2104-\allowbreak 2-\allowbreak i}, whereas Fashion-MNIST and CIFAR-10 target the Xilinx Zynq-7000 \texttt{XC7Z045-FFG900-2} to align with the state-of-the-art research. We synthesize the generated Verilog in Vivado 2025.2 using out-of-context synthesis with \nolinkurl{Flow_PerfOptimized_high}, followed by placement and routing. All DiffLUT-Net resource and timing values are post-implementation; baseline values are from their sources and are most directly comparable. 

Ours ($N$) in the result tables denotes one trainable LUT6 layer with N nodes, and Ours $(N_1, N_2)$ denotes two layers with $N_1$ and $N_2$ nodes. These are pre-implementation LUT numbers, not physical LUT numbers. The implementation number also includes GroupSum, while LUTs made unreachable by the learned mapping are removed. All reported DiffLUT-Net designs use one pipeline stage. We use $A\times L$ to denote the product of implemented LUT count and post-implementation latency. The LUT, flip-flop (FF), DSP, BRAM, and timing results reported for DiffLUT-Net are obtained from our own post-implementation reports. For gate-level models that do not provide FPGA LUT results, the \textit{Gates} column reports the source-reported number of trainable logic gates; this column is populated only for gate-level models.

\subsection{Results on JSC Benchmarks}

Table~\ref{tab:fpga-benchmarks} compares DiffLUT-Net with previous JSC implementations at compact configurations and high-accuracy configurations.

\begin{table*}[t]
  \caption{Classification accuracy, FPGA resource utilization, and latency comparison on JSC CERNBox and JSC OpenML. 
  }
  \label{tab:fpga-benchmarks}
  \centering
  \setlength{\tabcolsep}{2.2pt}
  \begin{tabular*}{\textwidth}{@{\extracolsep{\fill}}lrrrrrrr@{}}
    \toprule
    Method & Acc. (\%)~$\uparrow$ & LUTs~$\downarrow$ & FFs~$\downarrow$ & DSPs~$\downarrow$ & BRAMs~$\downarrow$ & Lat. (ns)~$\downarrow$ & $A\times L$~$\downarrow$ (LUT$\cdot$ns) \\
    \midrule
    \multicolumn{8}{l}{\textit{JSC CERNBox, low-accuracy}} \\
    LogicNets~\cite{Umuroglu2020LogicNets} & 71.8 & 37,931 & 810 & 0 & 0 & 13.0 & $4.93\times10^{5}$ \\
    PolyLUT~\cite{Andronic2023PolyLUT} & 72.0 & 12,436 & 773 & 0 & 0 & 5.0 & $6.20\times10^{4}$ \\
    NeuraLUT~\cite{Andronic2024NeuraLUT} & 72.0 & 4,684 & 341 & 0 & 0 & 3.00 & $1.40\times10^{4}$ \\
    \textbf{Ours (50)} & \textbf{72.5} & \textbf{94} & \textbf{20} & 0 & 0 & \textbf{1.21} & $\mathbf{1.13\times10^{2}}$ \\
    \midrule
    \multicolumn{8}{l}{\textit{JSC CERNBox, high-accuracy}} \\
    PolyLUT~\cite{Andronic2023PolyLUT} & 75.0 & 246,071 & 12,384 & 0 & 0 & 25.0 & $6.15\times10^{6}$ \\
    NeuraLUT~\cite{Andronic2024NeuraLUT} & \textbf{75.1} & 92,357 & 4,885 & 0 & 0 & 14.0 & $1.29\times10^{6}$ \\
    PolyLUT-Add~\cite{Lou2024PolyLUTAdd} & 75.0 & 36,484 & 1,209 & 0 & 0 & 16.0 & $5.84\times10^{5}$ \\
    AmigoLUT-NeuraLUT~\cite{Weng2025AmigoLUT} & 74.4 & 42,742 & 4,717 & 0 & 0 & 9.6 & $4.10\times10^{5}$ \\
    FPGN~\cite{Liang2026FPGN} & 74.9 & 12,358 & 4,839 & 0 & 0 & 6.0 & $7.41\times10^{4}$ \\
    NeuraLUT-Assemble~\cite{Andronic2025NeuraLUTAssemble} & 75.0 & 8,539 & 1,332 & 0 & 0 & 5.7 & $4.87\times10^{4}$ \\
    KANEL\'E~\cite{Hoang2026KANELE} & \textbf{75.1} & \textbf{5,034} & 1,917 & 0 & 0 & 8.1 & $4.10\times10^{4}$ \\
    \textbf{Ours (4000, 2000)} & 75.0 & 5,910 & \textbf{45} & 0 & 0 & \textbf{5.4} & $\mathbf{3.19\times10^{4}}$ \\
    \midrule
    \multicolumn{8}{l}{\textit{JSC OpenML, low-accuracy}} \\
    DWN ($n=6$, sm.)~\cite{Bacellar2024DWN} & 71.1 & 20 & 22 & 0 & 0 & 0.60 & $1.30\times10^{1}$ \\
    \textbf{Ours (10)} & 71.5 & \textbf{19} & \textbf{10} & 0 & 0 & \textbf{0.58} & $\mathbf{1.09\times10^{1}}$ \\
    DWN ($n=6$, sm.)~\cite{Bacellar2024DWN} & 74.0 & 110 & 72 & 0 & 0 & 1.50 & $2.00\times10^{2}$ \\
    \textbf{Ours (50)} & 74.1 & \textbf{94} & \textbf{20} & 0 & 0 & \textbf{1.21} & $\mathbf{1.14\times10^{2}}$ \\
    \midrule
    \multicolumn{8}{l}{\textit{JSC OpenML, high-accuracy}} \\
    hls4ml (Fahim et al.)~\cite{Fahim2021HLS4ML} & 76.2 & 63,251 & 4,394 & 38 & 0 & 45 & $2.85\times10^{6}$ \\
    da4ml~\cite{Sun2025DA4ML} & \textbf{76.9} & 12,250 & 1,502 & 0 & 0 & 18.9 & $2.30\times10^{5}$ \\
    DWN~\cite{Bacellar2024DWN} & 76.3 & 4,972 & 3,305 & 0 & 0 & 7.3 & $3.60\times10^{4}$ \\
    FPGN~\cite{Liang2026FPGN} & 76.0 & 3,345 & 1,703 & 0 & 0 & 5.5 & $1.84\times10^{4}$ \\
    KANEL\'E~\cite{Hoang2026KANELE} & 76.0 & \textbf{1,232} & 900 & 0 & 0 & 7.1 & $8.70\times10^{3}$ \\
    TreeLUT~\cite{Khataei2025TreeLUT} & 75.6 & 2,234 & 347 & 0 & 0 & 2.70 & $6.03\times10^{3}$ \\
    NeuraLUT-Assemble~\cite{Andronic2025NeuraLUTAssemble} & 76.0 & 1,780 & 540 & 0 & 0 & \textbf{2.10} & $\mathbf{3.92\times10^{3}}$ \\
    \textbf{Ours (2000, 500)} & 76.0 & 1,724 & \textbf{35} & 0 & 0 & 3.86 & $6.65\times10^{3}$ \\
    \bottomrule
  \end{tabular*}
  \parbox{\textwidth}{\footnotesize\raggedright
    \textit{Notes:} \texttt{--} indicates a value not reported by the corresponding source.
    $A\times L$ is the product of implemented LUT count $A$ and post-implementation latency $L$.
  }
\end{table*}

\paragraph{JSC CERNBox}
At the compact configuration, \textbf{Ours (50)} achieves 72.5\% accuracy using 94 LUTs, 20 FFs, and no DSPs or BRAMs, with 1.21\, ns latency. Compared with NeuraLUT at a similar accuracy of 72.0\%, it uses $49.8\times$ fewer LUTs and reduces $A\times L$ from $1.40\times10^{4}$ to $1.13\times10^{2}$ LUT$\cdot$ns, a reduction of approximately $124\times$. LogicNets and PolyLUT require substantially more LUTs while attaining slightly lower accuracy. 

At the high-accuracy configurations, \textbf{Ours (4000, 2000)} reaches 75.0\% accuracy with \pslutCernFourKTwoK{} post-synthesis LUTs in the first and the second layer, respectively. This LUT-network requires
5910 post-implementation LUTs, 45 FFs, leading to  5.4\, ns latency. It matches the accuracy of NeuraLUT-Assemble while using 31\% fewer post-implementation LUTs and reduces $A\times L$ by 35\%. KANEL\'E reaches 75.1\% accuracy with 15\% fewer LUTs, whereas DiffLUT-Net reduces its latency by 33\% and its $A\times L$ by 22\%. As a result, \textbf{Ours (4000, 2000)} achieves the lowest $A\times L$ among the high-accuracy configurations for CERNBox in Table~\ref{tab:fpga-benchmarks}.

\paragraph{JSC OpenML}
The compact configurations for JSC OpenML provide direct resource-matched comparisons with DWN. \textbf{Ours (10)} improves accuracy from 71.1\% to 71.5\% while reducing the LUT numbers from 20 to 19, latency from 0.60 to 0.58\, ns, and $A\times L$ from 13.0 to 10.9 LUT$\cdot$ns. At the second compact configuration, \textbf{Ours (50)} reaches 74.1\% accuracy, compared with 74.0\% for DWN, while reducing LUT numbers from 110 to 94, latency from 1.50 to 1.21\, ns, and $A\times L$ from 200 to 114 LUT$\cdot$ns. The improvement at both model scales shows that the learned LUT functions and connections remain effective even in networks containing only a small number of trainable LUT6s.

At the high-accuracy configuration,  with 76.0\% accuracy, \textbf{Ours (2000, 500)} has \pslutOpenmlTwoKFiveHundred{} post-synthesis LUTs for the first and the second LUT layer, respectively. It uses 1,724 post-implementation LUTs and 35 FFs with a latency of 3.86\, ns. KANEL\'E uses fewer LUTs, but DiffLUT-Net reduces its latency by 46\% and its $A\times L$ by 24\%. NeuraLUT-Assemble achieves lower latency and $A\times L$, whereas DiffLUT-Net uses 3\% fewer LUTs. NeuraLUT-Assemble's latency advantage largely comes from its architecture, which does not require the GroupSum adder tree used by DiffLUT-Net.

Across the two JSC benchmarks, DiffLUT-Net achieves a good accuracy--hardware trade-off in both compact and high-accuracy configurations. The compact configurations provide strong resource efficiency at comparable accuracy, while the high-accuracy configurations remain competitive as the target accuracy increases.

\subsection{Results on MNIST}

Table~\ref{tab:mnist} reports the performance of DiffLUT-Net on MNIST with and without data augmentation, together with a high-accuracy configuration used to evaluate the attainable classification accuracy.

\begin{table*}[t]
  \caption{Classification accuracy, FPGA resource utilization, and latency comparison on MNIST.}
  \label{tab:mnist}
  \centering
  \setlength{\tabcolsep}{1.3pt}
  \begin{tabular*}{\textwidth}{@{\extracolsep{\fill}}lrrrrrrrr@{}}
    \toprule
    Method
    & Acc. (\%)~$\uparrow$
    & Gates
    & LUTs~$\downarrow$
    & FFs~$\downarrow$
    & DSPs~$\downarrow$
    & BRAMs~$\downarrow$
    & Lat. (ns)~$\downarrow$
    & $A\times L$~$\downarrow$ (LUT$\cdot$ns)
    \\
    \midrule
    \multicolumn{9}{l}{\textit{Arithmetic neural-network accelerators (mixed FPGA targets)}} \\
    hls4ml (Ngadiuba et al.)~\cite{Ngadiuba2021HLS4MLCompression}
    & 95.0 & -- & 260,092 & 165,513 & 0 & 345 & 190 & $4.94\times10^{7}$ \\
    FINN~\cite{Umuroglu2017FINN}
    & 96.0 & -- & 91,131 & -- & 0 & 5 & 310.0 & $2.82\times10^{7}$ \\
    \midrule
    \multicolumn{9}{l}{\textit{Post-training truth-table compilation} (\texttt{xcvu9p-flgb2104-2-i})} \\
    PolyLUT~\cite{Andronic2023PolyLUT}
    & 97.5 & -- & 75,131 & 4,668 & 0 & 0 & 17.0 & $1.28\times10^{6}$ \\
    NeuraLUT~\cite{Andronic2024NeuraLUT}
    & 96.0 & -- & 54,798 & 3,757 & 0 & 0 & 12.0 & $6.58\times10^{5}$ \\
    PolyLUT-Add~\cite{Lou2024PolyLUTAdd}
    & 96.0 & -- & 14,810 & 2,609 & 0 & 0 & 10.0 & $1.48\times10^{5}$ \\
    AmigoLUT-NeuraLUT~\cite{Weng2025AmigoLUT}
    & 95.5 & -- & 16,081 & 13,292 & 0 & 0 & 7.6 & $1.22\times10^{5}$ \\
    KANEL\'E~\cite{Hoang2026KANELE}
    & 96.3 & -- & 3,809 & 4,133 & 0 & 0 & 9.3 & $3.50\times10^{4}$ \\
    TreeLUT~\cite{Khataei2025TreeLUT}
    & 96.6 & -- & 4,478 & 597 & 0 & 0 & 2.50 & $1.12\times10^{4}$ \\
    NeuraLUT-Assemble$^{*}$~\cite{Andronic2025NeuraLUTAssemble}
    & 98.6 & -- & 5,037 & 713 & 0 & 0 & 2.20 & $1.11\times10^{4}$ \\
    NeuraLUT-Assemble~\cite{Andronic2025NeuraLUTAssemble}
    & 97.9 & -- & 5,070 & 725 & 0 & 0 & \textbf{2.10} & $\mathbf{1.06\times10^{4}}$ \\
    \midrule
    \multicolumn{9}{l}{\textit{Directly trained logic and LUT networks (mixed FPGA targets)}} \\
    DiffLogic~\cite{Petersen2022DiffLogic}
    & 98.5 & 384K & -- & -- & -- & -- & -- & -- \\
    LILogicNet$^{*}$~\cite{Fojcik2025LILogicNet}
    & 99.0 & 32K & 37,373 & -- & -- & -- & -- & -- \\
    LILogicNet$^{*}$~\cite{Fojcik2025LILogicNet}
    & 98.5 & 8K & 14,076 & -- & -- & -- & -- & -- \\
    LILogicNet$^{*}$~\cite{Fojcik2025LILogicNet}
    & 98.0 & 4K & 7,103 & -- & -- & -- & -- & -- \\
    DWN~\cite{Bacellar2024DWN}
    & 98.3 & -- & 4,082 & 3,385 & 0 & 0 & 6.0 & $2.40\times10^{4}$ \\
    LGN~\cite{Ma2026CLGN}
    & 98.2 & -- & 7,768 & -- & -- & -- & 8.5 & -- \\
    \midrule
    \multicolumn{9}{l}{\textit{DiffLUT-Net} (\texttt{xcvu9p-flgb2104-2-i})} \\
    \textbf{Ours (2000)}$^{*}$
    & 98.6 & -- & 4,669 & 80 & 0 & 0 & 3.41 & $1.59\times10^{4}$ \\
    \textbf{Ours (2000, 1000)}
    & 98.0 & -- & \textbf{2,990} & \textbf{70} & 0 & 0 & 4.11 & $1.23\times10^{4}$ \\
    \textbf{Ours (8000)}$^{*}$
    & \textbf{99.3} & -- & 18,764 & 100 & 0 & 0 & 4.46 & $8.37\times10^{4}$ \\
    \bottomrule
  \end{tabular*}
  \parbox{\textwidth}{\footnotesize\raggedright
    \textit{Notes:} $^{*}$ indicates data augmentation, and \texttt{--} indicates a value not reported by the corresponding source.
    \textit{Gates} is populated only for gate-level models and denotes the source-reported number of trainable logic gates, not an implemented FPGA resource count.
    $A\times L$ is the product of implemented LUT count $A$ and reported latency $L$.
  }
\end{table*}

Without data augmentation, \textbf{Ours (2000, 1000)} has \pslutMnistTwoKOneK{} post-synthesis LUTs for the first and the second LUT layers, respectively. It achieves 98.0\% accuracy, 0.1\% higher than NeuraLUT-Assemble, while using 41\% fewer post-implementation LUTs. With data augmentation, \textbf{Ours (2000)} matches the 98.6\% accuracy of NeuraLUT-Assemble while using 7\% fewer LUTs. These results show that DiffLUT-Net achieves good accuracy--resource trade-offs under both training settings.

In the high-accuracy configuration, \textbf{Ours (8000)} reaches 99.3\% accuracy, the highest value reported in Table~\ref{tab:mnist}, while using approximately half as many LUTs as the 99.0\%-accurate LILogicNet configuration. This result demonstrates that DiffLUT-Net can also be scaled to achieve high classification accuracy with competitive FPGA resource utilization.

\subsection{Results on Fashion-MNIST and CIFAR-10}

Table~\ref{tab:fashion-cifar} reports the accuracy and logic-resource results on Fashion-MNIST and CIFAR-10.

\begin{table}[t]
  \caption{Comparison of accuracy and resource utilization on Fashion-MNIST and CIFAR-10 for directly trained logic and LUT networks.}
  \label{tab:fashion-cifar}
  \centering
  \setlength{\tabcolsep}{1.5pt}
  \begin{tabular*}{\columnwidth}{@{\extracolsep{\fill}}lrrr@{}}
    \toprule
    Method & Acc. (\%)~$\uparrow$ & Gates & LUTs~$\downarrow$ \\
    \midrule
    \multicolumn{4}{l}{\textit{Fashion-MNIST}} \\
    LILogicNet~\cite{Fojcik2025LILogicNet} & 90.61 & 128K & 143,226 \\
    LILogicNet~\cite{Fojcik2025LILogicNet} & 90.26 & 64K & 73,938 \\
    LILogicNet~\cite{Fojcik2025LILogicNet} & 89.95 & 8K & 14,321 \\
    DWN~\cite{Bacellar2024DWN} & 89.01 & -- & 6,200 \\
    \textbf{Ours (2000, 1000)} & 89.00 & -- & \textbf{2,994} \\
    \textbf{Ours (8000)} & 90.40 & -- & 18,748 \\
    \midrule
    \multicolumn{4}{l}{\textit{CIFAR-10}} \\
    DiffLogic~\cite{Petersen2022DiffLogic} & 57.39 & 512K & -- \\
    LILogicNet$^{*}$~\cite{Fojcik2025LILogicNet} & \textbf{60.98} & 256K & 293,285 \\
    LILogicNet$^{*}$~\cite{Fojcik2025LILogicNet} & 57.66 & 64K & 104,853 \\
    LILogicNet$^{*}$~\cite{Fojcik2025LILogicNet} & 55.11 & 8K & 14,415 \\
    LGN(+)~\cite{Ma2026CLGN} & 58.85 & -- & 79,375 \\
    LGN~\cite{Ma2026CLGN} & 57.71 & -- & 38,946 \\
    DWN~\cite{Bacellar2024DWN} & 57.42 & -- & 16,700 \\
    FPGN~\cite{Liang2026FPGN} & 58.10 & -- & 15,336 \\
    \textbf{Ours (6000)}$^{*}$ & 59.50 & -- & \textbf{13,951} \\
    \bottomrule
  \end{tabular*}
  \parbox{\columnwidth}{\footnotesize\raggedright
    \textit{Notes:} $^{*}$ indicates data augmentation, and \texttt{--} indicates a value not reported by the corresponding source.
    \textit{Gates} is populated only for gate-level models and denotes the source-reported number of trainable logic gates.
    DiffLUT-Net and DWN target the same \texttt{XC7Z045-FFG900-2} FPGA part.
  }
\end{table}

\paragraph{Fashion-MNIST}
The compact \textbf{Ours (2000, 1000)} configuration has \pslutFashionTwoKOneK{} post-synthesis LUTs in the first and the second layers, respectively. It achieves 89.00\% accuracy, closely matching the 89.01\% accuracy of DWN while reducing the post-implementation LUT count from 6200 to 2994. This corresponds to a 52\% reduction and is the smallest reported LUT count in the Fashion-MNIST comparison. In the high-accuracy configuration, \textbf{Ours (8000)}  reaches 90.40\% accuracy using 18748 LUTs. Compared with the 90.26\%-accurate LILogicNet configuration, it improves accuracy by 0.14\% while using 75\% fewer LUTs. It also comes within 0.21\% of the highest LILogicNet result of 90.61\%, while using $7.6\times$ fewer LUTs. These results demonstrate good accuracy--LUT trade-offs for both compact and higher-accuracy DiffLUT-Net configurations.

\paragraph{CIFAR-10}
With data augmentation, \textbf{Ours (6000)} achieves 59.50\% accuracy using 13951 LUTs, the smallest reported FPGA LUT count in the CIFAR-10 comparison. Compared with DWN, it improves accuracy by 2.08\% while using 16\% fewer LUTs. It also exceeds FPGN by 1.40\% while using 9\% fewer LUTs. The highest baseline accuracy is 60.98\%, only 1.48\% higher than DiffLUT-Net, but requires $21.0\times$ more LUTs.

Overall, DiffLUT-Net achieves the best accuracy--LUT trade-offs on both datasets, matching or improving the closest baselines with substantially lower FPGA resource utilization.

\subsection{Ablation Study}

We conduct four ablation studies to examine how the main design choices of DiffLUT-Net affect classification accuracy and FPGA implementation cost. The experiments study the width of a single LUT6 layer, network depth, the GroupSum training temperature, and the width of a second LUT6 layer.

\subsubsection{Influence of the Single-Layer Width}

We first vary the width $N$ of a single-layer DiffLUT-Net, where $N$ denotes the number of trainable LUTs before implementation.

\begin{figure}[t]
  \centering
  \resizebox{\linewidth}{!}{%
  \begingroup
\definecolor{linegray}{HTML}{98A2B3}
\definecolor{gridgray}{HTML}{D9DEE7}

\begin{tikzpicture}
\begin{axis}[
  width=0.92\linewidth,height=0.57\linewidth,
  xmin=0,xmax=10000,
  ymin=73.8,ymax=76.45,
  xtick={0,2000,4000,6000,8000,10000},
  every x tick scale label/.append style={xshift=20pt,yshift=2pt},
  xlabel={Post-implementation LUTs},
  ylabel={Accuracy (\%)},
  tick label style={font=\footnotesize},
  label style={font=\small},
  axis line style={line width=0.7pt},
  axis lines*=left,
  grid=major,
  grid style={gridgray,line width=0.45pt},
  colorbar,
  colormap/viridis,
  point meta min=1.5,
  point meta max=4.73,
  colorbar style={width=3mm,ylabel={Latency (ns)},ylabel style={font=\footnotesize,xshift=-1pt},tick label style={font=\scriptsize}},
  clip=false,
]
\addplot[linegray,line width=0.8pt] coordinates {
  (94,74.1) (193,74.8) (555,75.3) (1225,75.7)
  (2333,75.9) (4653,76.1) (9379,76.3)
};
\addplot[
  only marks,scatter,scatter src=explicit,
  mark=*,draw=white,line width=0.45pt,
  visualization depends on={\thisrow{ms}\as\perpointmarksize},
  scatter/@pre marker code/.append style={/tikz/mark size=\perpointmarksize}
] table[x=lut,y=acc,meta=latency] {
lut acc latency ms
94  74.1 1.21 1.5
193  74.8 1.51 1.8
555  75.3 2.39 2.1
1225 75.7 3.00 2.4
2333 75.9 3.28 2.7
4653 76.1 4.425 3.0
9379 76.3 4.73 3.3
};
\node[font=\footnotesize,anchor=south west,inner sep=1pt,fill=white,fill opacity=0.82,text opacity=1,yshift=2pt] at (axis cs:180,74.02) {$N=50$};
\node[font=\footnotesize,anchor=south west,inner sep=1pt,fill=white,fill opacity=0.82,text opacity=1,yshift=2pt] at (axis cs:320,74.62) {$N=100$};
\node[font=\footnotesize,anchor=south west,inner sep=1pt,fill=white,fill opacity=0.82,text opacity=1,yshift=2pt] at (axis cs:760,75.12) {$N=250$};
\node[font=\footnotesize,anchor=south west,inner sep=1pt,fill=white,fill opacity=0.82,text opacity=1,yshift=2pt] at (axis cs:530,75.78) {$N=540$};
\node[font=\footnotesize,anchor=north west,inner sep=1pt,fill=white,fill opacity=0.82,text opacity=1,yshift=-2pt] at (axis cs:2050,76.32) {$N=1000$};
\node[font=\footnotesize,anchor=south west,inner sep=1pt,fill=white,fill opacity=0.82,text opacity=1,yshift=2pt] at (axis cs:4850,76.12) {$N=2000$};
\node[font=\footnotesize,anchor=north east,inner sep=1pt,fill=white,fill opacity=0.82,text opacity=1,yshift=-2pt] at (axis cs:9850,76.25) {$N=4000$};

\node[font=\small,anchor=west] at (axis cs:8350,74.68) {FFs};
\draw[fill=gray!75,draw=white] (axis cs:8380,74.48) circle[radius=1.3pt];
\node[font=\footnotesize,anchor=west] at (axis cs:8740,74.48) {20};
\draw[fill=gray!75,draw=white] (axis cs:8380,74.28) circle[radius=2.0pt];
\node[font=\footnotesize,anchor=west] at (axis cs:8740,74.28) {35};
\draw[fill=gray!75,draw=white] (axis cs:8380,74.08) circle[radius=2.7pt];
\node[font=\footnotesize,anchor=west] at (axis cs:8740,74.08) {50};
\end{axis}
\end{tikzpicture}
\endgroup%
}
  \caption{Accuracy--hardware trade-off on JSC OpenML as the width $N$ of a single LUT6 layer increases from 50 to 4,000. The x-axis reports post-implementation LUT utilization; the y-axis reports classification accuracy. Marker color denotes latency, and marker area denotes FF utilization.}
  \label{fig:openml_scale_ablation}
  \Description{Scatter plot of JSC OpenML accuracy against post-implementation LUT utilization for different single-layer widths. Marker color represents latency, and marker area represents FF utilization.}
\end{figure}
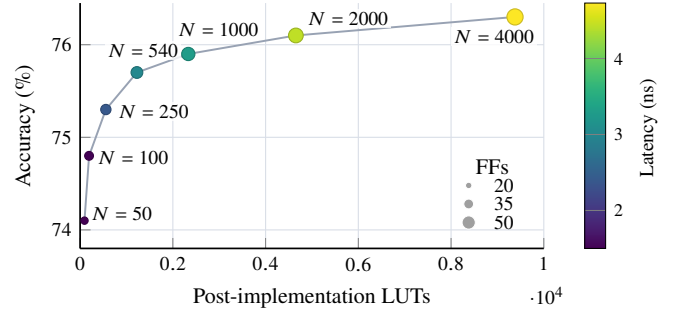

Figure~\ref{fig:openml_scale_ablation} shows that increasing $N$ consistently improves accuracy, but with diminishing benefits. Scaling from $N=50$ to $N=4000$ increases accuracy from 74.1\% to 76.3\%. The diminishing benefit is clearest beyond $N=1000$: increasing the width fourfold from 1000 to 4000 improves accuracy by only 0.4\%. Viewed on a logarithmic scale, accuracy increases approximately linearly with model width, exhibiting a scaling-law-like trend; on the linear hardware scale, however, this corresponds to diminishing accuracy gains per additional LUT.

\subsubsection{Influence of Network Depth}
\label{sec:network_depth}

We fix every LUT6 layer with 2,000 trainable LUTs and vary the network depth from one to four layers. This experiment isolates the effect of stacking additional learned LUT layers while keeping the nominal width of each layer unchanged. 

\begin{figure}[t]
  \centering
  \resizebox{\linewidth}{!}{%
  \begingroup
\definecolor{linegray}{HTML}{98A2B3}
\definecolor{gridgray}{HTML}{D9DEE7}

\begin{tikzpicture}
\begin{axis}[
  width=0.92\linewidth,height=0.57\linewidth,
  xmin=4000,xmax=10000,
  ymin=75.6,ymax=76.7,
  xtick={0,2000,4000,6000,8000,10000,12000},
  every x tick scale label/.append style={xshift=20pt,yshift=2pt},
  xlabel={Post-implementation LUTs},
  ylabel={Accuracy (\%)},
  tick label style={font=\footnotesize},
  label style={font=\small},
  axis line style={line width=0.7pt},
  axis lines*=left,
  grid=major,
  grid style={gridgray,line width=0.45pt},
  colorbar,
  colormap/viridis,
  point meta min=4,
  point meta max=10,
  colorbar style={
    width=3mm,
    ylabel={Latency (ns)},
    ylabel style={font=\footnotesize,xshift=-1pt},
    tick label style={font=\scriptsize}
  },
  clip=false,
]

\addplot[linegray,line width=0.8pt] coordinates {
  (4653,76.1)
  (5542,76.5)
  (6226,76.3)
  (7961,75.8)
};

\addplot[
  only marks,
  scatter,
  scatter src=explicit,
  mark=*,
  mark size=3.2pt,
  draw=white,
  line width=0.45pt
] table[x=lut,y=acc,meta=latency] {
  lut   acc   latency
  4653  76.1  4.425
  5542  76.5  5.530
  6226  76.3  6.431
  7961  75.8  6.922
};

\node[
  font=\footnotesize,
  anchor=south east,
  inner sep=1pt,
  fill=white,
  fill opacity=0.82,
  text opacity=1,
  yshift=2pt
] at (axis cs:4653,76.1) {$1$};

\node[
  font=\footnotesize,
  anchor=south west,
  inner sep=1pt,
  fill=white,
  fill opacity=0.82,
  text opacity=1,
  yshift=2pt
] at (axis cs:5542,76.5) {$2$};

\node[
  font=\footnotesize,
  anchor=south west,
  inner sep=1pt,
  fill=white,
  fill opacity=0.82,
  text opacity=1,
  yshift=2pt
] at (axis cs:6226,76.3) {$3$};

\node[
  font=\footnotesize,
  anchor=north west,
  inner sep=1pt,
  fill=white,
  fill opacity=0.82,
  text opacity=1,
  yshift=-2pt
] at (axis cs:7961,75.8) {$4$};

\node[font=\footnotesize,anchor=south west]
  at (rel axis cs:0.03,0.03) {Labels: number of layers};
\end{axis}
\end{tikzpicture}
\endgroup%
}
  \caption{Effect of network depth on JSC OpenML when every LUT6 layer contains 2000 trainable LUTs. The x-axis reports post-implementation LUT utilization, the y-axis reports classification accuracy, and marker color denotes post-implementation latency. The labels beside the markers denote the number of LUT6 layers.}
  \label{fig:openml_depth_ablation}
  \Description{Scatter plot of JSC OpenML accuracy against post-implementation LUT utilization for networks containing one to four 2,000-node LUT6 layers. Marker color represents latency, and the marker labels identify the number of layers.}
\end{figure}
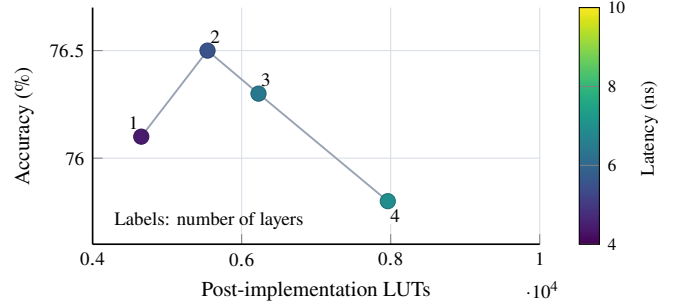
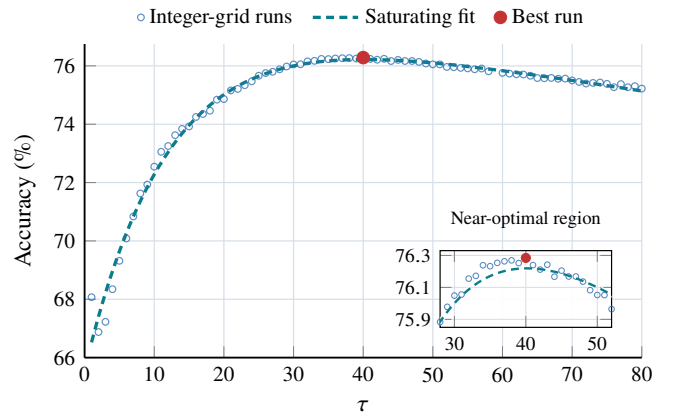
\begin{figure}[!tbp]
  \centering
  \begingroup
\definecolor{trend}{HTML}{087E8B}
\definecolor{points}{HTML}{2F6FB0}
\definecolor{bestpoint}{HTML}{C43333}
\definecolor{gridgray}{HTML}{D6DEE8}

\begin{tikzpicture}
\begin{axis}[
  name=mainaxis,
  width=\linewidth,height=0.64\linewidth,
  xmin=0,xmax=80,ymin=66,ymax=76.75,
  xtick={0,10,...,80},ytick={66,68,...,76},
  xlabel={$\tau$},ylabel={Accuracy (\%)},
  tick label style={font=\footnotesize},label style={font=\small},
  axis lines*=left,
  axis line style={line width=0.7pt},
  grid=major,grid style={gridgray,line width=0.45pt},
  legend style={draw=none,font=\footnotesize,at={(0.5,1.02)},anchor=south,legend columns=3,
    /tikz/every even column/.append style={column sep=5pt}},
  clip=false,
]
\addplot[only marks,mark=o,mark size=1.25pt,draw=points,fill=white,
  line width=0.45pt,opacity=0.78]
  table[x=tau,y=accuracy,col sep=comma] {figures/latex_comparison/wandb_tau_accuracy.csv};
\addlegendentry{Integer-grid runs}
\addplot[white,densely dashed,line width=2.7pt,opacity=0.90,forget plot]
  table[x=tau,y=accuracy,col sep=comma] {figures/latex_comparison/wandb_tau_accuracy_fit.csv};
\addplot[trend,densely dashed,line width=1.25pt]
  table[x=tau,y=accuracy,col sep=comma] {figures/latex_comparison/wandb_tau_accuracy_fit.csv};
\addlegendentry{Saturating fit}
\addplot[only marks,mark=*,mark size=2.35pt,bestpoint]
  coordinates {(40,76.28433734939759)};
\addlegendentry{Best run}
\end{axis}

\begin{axis}[
  at={(mainaxis.south east)},anchor=south east,
  xshift=-4mm,yshift=4mm,
  width=0.43\linewidth,height=0.29\linewidth,
  xmin=28,xmax=52,ymin=75.85,ymax=76.33,
  xtick={30,40,50},ytick={75.9,76.1,76.3},
  tick label style={font=\scriptsize},
  title={Near-optimal region},title style={font=\scriptsize,yshift=-1pt},
  axis lines=box,axis line style={line width=0.45pt},
  grid=major,grid style={gridgray,line width=0.25pt},
  axis background/.style={fill=white,fill opacity=0.95},
]
\addplot[only marks,mark=o,mark size=1.05pt,draw=points,fill=white,
  line width=0.4pt,opacity=0.82]
  table[x=tau,y=accuracy,col sep=comma] {figures/latex_comparison/wandb_tau_accuracy.csv};
\addplot[white,densely dashed,line width=1.9pt,opacity=0.90,forget plot]
  table[x=tau,y=accuracy,col sep=comma] {figures/latex_comparison/wandb_tau_accuracy_fit.csv};
\addplot[trend,densely dashed,line width=0.9pt]
  table[x=tau,y=accuracy,col sep=comma] {figures/latex_comparison/wandb_tau_accuracy_fit.csv};
\addplot[only marks,mark=*,mark size=1.7pt,bestpoint]
  coordinates {(40,76.28433734939759)};
\end{axis}
\end{tikzpicture}
\endgroup%
  \caption{Effect of the GroupSum training temperature $\tau$ on JSC OpenML accuracy for a two-layer DiffLUT-Net with 1000 and 500 trainable LUT6s. Open circles denote completed runs, the dashed curve shows the fitted trend, and the highlighted marker identifies the best observed run at $\tau=40$ with 76.3\% accuracy.}
  \label{fig:tau_accuracy_sweep}
  \Description{Scatter plot of JSC OpenML accuracy across the GroupSum temperature sweep, with a fitted trend and the best observed run highlighted at tau 40.}
\end{figure}

\begin{figure*}[!t]
  \centering
  \includegraphics[width=\textwidth]{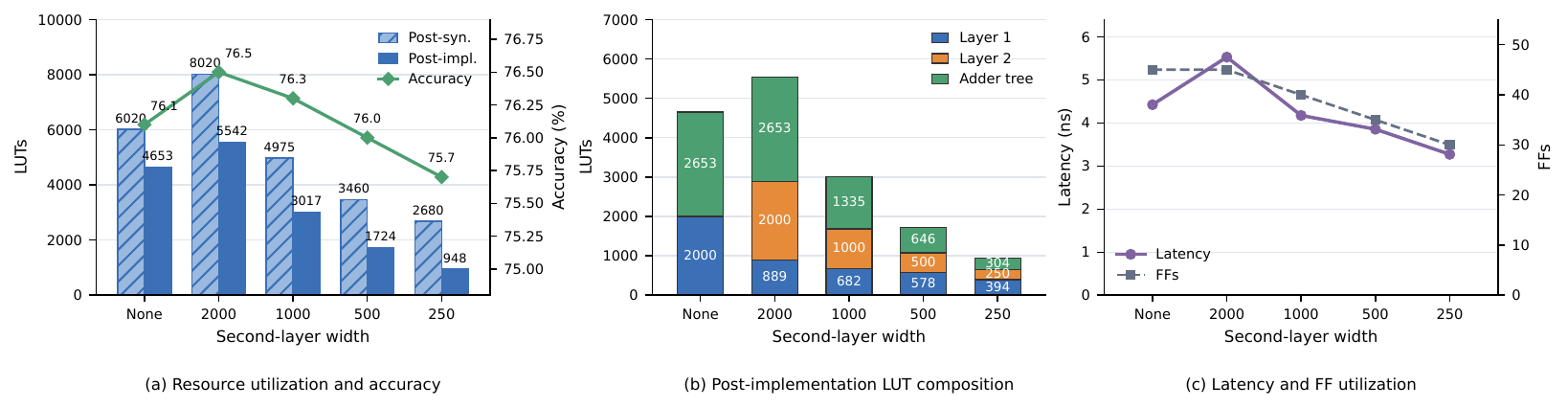}
  \caption{Effect of the second-layer width with the first LUT6 layer fixed at 2000 LUTs. \texttt{None} denotes the single-layer reference. (a) Post-synthesis and post-implementation total LUT utilization together with classification accuracy. (b) Post-implementation LUT composition of the retained first-layer logic, the second layer, and the GroupSum adder tree. (c) Post-implementation latency and FF utilization.}
  \label{fig:openml_ablation_three_corrected}
  \Description{Three-panel ablation study of different second-layer widths with the first layer fixed at 2,000 nodes. The panels report total LUT utilization and accuracy, post-implementation LUT utilization by architectural component, and post-implementation latency and FF utilization.}
\end{figure*}
Figure~\ref{fig:openml_depth_ablation} shows that increasing depth does not produce a monotonic accuracy improvement. Accuracy rises from 76.1\% with one layer to 76.5\% with two layers, but then decreases to 76.3\% with three layers and 75.8\% with four layers. Among the evaluated configurations, the two-layer network therefore provides the highest classification accuracy. We have examined the structure of the four-layer LUT network. It is 
implemented with 1837, 1061, 435, and 1980 LUTs in its four successive LUT6 layers. In particular, the third layer is reduced to only 435 LUTs after training. This indicates that the learned connections in the final layer use a relatively small subset of upstream signals, forming a narrow bottleneck before the output stage. The additional layers therefore increase hardware cost without providing more information, which helps explain the four-layer model's lower accuracy. Similar degradation with increasing depth has also been observed in differentiable logic networks, where deeper models are more difficult to optimize and more sensitive to discretization errors~\cite{Ruttgers2025Light}.

\subsubsection{Influence of the GroupSum Training Temperature $\tau$}

Following the GroupSum formulation in Section~\ref{sec:grouped_output}, we vary the training temperature $\tau$ to evaluate its effect on optimization. All runs use a two-layer DiffLUT-Net with 1000 and 500 trainable LUTs in the first and second layers, respectively.

Figure~\ref{fig:tau_accuracy_sweep} shows an approximately inverted-U-shaped relationship between $\tau$ and classification accuracy. The best observed run reaches 76.3\% accuracy at $\tau=40$, and the strongest results are concentrated between approximately $\tau=34$ and $\tau=43$. Accuracy decreases toward both ends of the evaluated range, indicating that an intermediate temperature provides the most effective training behavior. 

\subsubsection{Influence of the Second-Layer Width}
\label{sec:second_layer_width}

The final ablation fixes the first LUT6 layer with 2000 LUTs and varies the second-layer width among 2000, 1000, 500, and 250 LUTs. The configuration labeled “None” is the single-layer baseline, in which the 2000 LUTs in the first layer connect directly to GroupSum. This experiment examines how the second-layer width affects not only the second layer itself, but also the number of LUTs before and after it.

To preserve the module boundaries required for the component-level breakdown, synthesis is performed with \texttt{KEEP\_HIERARCHY = yes}, while optimization remains enabled within each module. The total resource and timing results are taken after implementation.

Figure~\ref{fig:openml_ablation_three_corrected} shows that reducing the second-layer width produces three related hardware savings. First, the narrower second layer directly contains fewer LUTs. Second, it selects fewer outputs from the first layer, allowing first-layer LUTs with no remaining path to the network output to be removed during implementation. Third, the narrower final layer supplies fewer inputs to GroupSum and therefore reduces the size of the class-score adder tree. The learned sparse mapping consequently affects the implementation cost of the preceding layer, the current layer, and the output accumulation logic.

The single-layer reference reaches 76.1\% accuracy using 4653 post-implementation LUTs. Adding a 2000-node second layer increases accuracy to 76.5\%, but also increases LUT utilization to 5542. A 1000-LUT second layer achieves 76.3\% accuracy using 3017 LUTs, improving accuracy by 0.2\% while reducing LUT utilization by 35\% relative to the single-layer reference. With 500 second-layer LUTs, the network retains 76.0\% accuracy using only 1,724 LUTs, a 63\% reduction. Reducing the second layer further to 250 nodes lowers LUT utilization to 948, but also reduces accuracy to 75.7\%.

The 500--1000-node range therefore provides the best trade-off in this experiment. It preserves nearly all of the classification accuracy while substantially reducing the second layer, the retained first-layer logic, and the GroupSum adder tree. Overall, learned sparse connectivity allows a narrower second layer to preserve accuracy while substantially reducing the total hardware cost.

\section{Conclusion}

This work introduced DiffLUT-Net, a complete train-to-deployment framework for FPGA-native LUT6 networks. Instead of training an arithmetic model and later converting it into logic, DiffLUT-Net jointly learns LUT6 functions and sparse hardware-valid connections. Distribution-aware thermometer encoding converts real-valued inputs into binary signals, and GroupSum produces the final class scores. After training, the learned functions and connections are hardened and exported as synthesizable Verilog, keeping the training representation aligned with the deployed FPGA hardware.

Experiments on JSC CERNBox, JSC OpenML, MNIST, Fashion-MNIST, and CIFAR-10 show that DiffLUT-Net achieves strong accuracy--hardware trade-offs across different model scales. Compact configurations provide high resource efficiency, while high-accuracy configurations improve accuracy without excessive hardware growth. The ablation studies further show that model width provides a controllable path for scaling accuracy and that learned sparse connectivity can reduce logic across multiple stages of the implemented network. 

Future work includes extending differentiable LUT networks to convolutional and transformer-based architectures with cross-layer parameter sharing~\cite{wang2025basis}.

\section*{Acknowledgement}
This work is funded by the European Union - European Research Council (ERC) Starting Grant - Project-ID 101219243. Views and opinions expressed are however those of the author(s) only and do not necessarily reflect those of the European Union or the European Research Council. Neither the European Union nor the granting authority can be held responsible for them.

\printbibliography

\clearpage
\appendix

\section{Training Configurations}
\label{app:training_configurations}

This appendix summarizes the training configurations used in the benchmark and ablation experiments. Table~\ref{tab:app_shared_hparams} lists the hyperparameters shared by all experiments, while the following tables report configuration-specific settings. The dataset-specific values of $L$ and $K$ follow the settings in the main text.

\begin{table}[H]
  \caption{Training settings shared by the benchmark and ablation experiments.}
  \label{tab:app_shared_hparams}
  \centering
  \small
  \begin{tabular}{@{}ll@{}}
    \toprule
    Setting & Value \\
    \midrule
    Batch size (BS)                        & 100 \\
    Evaluation frequency (EF)              & 1,000 \\
    \texttt{--learning-rate}               & $1\times10^{-7}$ \\
    \texttt{--anneal-lr}                   & Enabled \\
    \texttt{--lr-end}                      & $1\times10^{-9}$ \\
    \texttt{--grad-factor}                 & 1.0 \\
    \texttt{--penalty}                     & 1.0 \\
    \texttt{--lamda\_init}                 & 10 \\
    \bottomrule
  \end{tabular}
\end{table}

\subsection{Configurations for the JSC Benchmarks}
\label{app:jsc_training_configurations}

Table~\ref{tab:app_jsc_hparams} lists the DiffLUT-Net configurations reported in Table~\ref{tab:fpga-benchmarks}. All JSC models are trained without data augmentation.

\begin{table}[H]
  \caption{Training configurations used for the JSC benchmark results in Table~\ref{tab:fpga-benchmarks}.}
  \label{tab:app_jsc_hparams}
  \centering
  \small
  \begin{tabular}{@{}llcc@{}}
    \toprule
    Dataset & Configuration & $\tau$ & Training steps \\
    \midrule
    JSC CERNBox & Ours (50)         & 2.7  & 200,000 \\
    JSC CERNBox & Ours (4000, 2000) & 74.3 & 200,000 \\
    \midrule
    JSC OpenML  & Ours (10)         & 1.9  & 200,000 \\
    JSC OpenML  & Ours (50)         & 3.6  & 200,000 \\
    JSC OpenML  & Ours (2000, 500)  & 23.7 & 200,000 \\
    \bottomrule
  \end{tabular}
\end{table}

\subsection{Configurations for MNIST}
\label{app:mnist_training_configurations}

Table~\ref{tab:app_mnist_hparams} lists the training configurations reported in Table~\ref{tab:mnist}. The two-layer model is trained without data augmentation, while data augmentation is enabled for the two single-layer configurations marked with an asterisk in the main benchmark table.

\begin{table}[H]
  \caption{Training configurations used for the MNIST results in Table~\ref{tab:mnist}.}
  \label{tab:app_mnist_hparams}
  \centering
  \small
  \begin{tabular}{@{}lccc@{}}
    \toprule
    Configuration & Augmentation & $\tau$ & Training steps \\
    \midrule
    Ours (2000, 1000) & No  & 7.0  & 200,000 \\
    Ours (2000)       & Yes & 8.3  & 200,000 \\
    Ours (8000)       & Yes & 15.6 & 200,000 \\
    \bottomrule
  \end{tabular}
\end{table}

\subsection{Configurations for Fashion-MNIST and CIFAR-10}
\label{app:fashion_cifar_training_configurations}

Table~\ref{tab:app_fashion_cifar_hparams} lists the DiffLUT-Net configurations reported in Table~\ref{tab:fashion-cifar}. The Fashion-MNIST configurations are trained without data augmentation. Data augmentation is enabled for CIFAR-10, which is trained for 400,000 steps.

\begin{table}[H]
  \caption{Training configurations used for the Fashion-MNIST and CIFAR-10 results in Table~\ref{tab:fashion-cifar}.}
  \label{tab:app_fashion_cifar_hparams}
  \centering
  \small
  \begin{tabular}{@{}llccc@{}}
    \toprule
    Dataset & Configuration & Aug. & $\tau$ & Training steps \\
    \midrule
    Fashion-MNIST & Ours (2000, 1000) & No  & 10.4 & 200,000 \\
    Fashion-MNIST & Ours (8000)       & No  & 17.7 & 200,000 \\
    CIFAR-10      & Ours (6000)       & Yes & 22.9 & 400,000 \\
    \bottomrule
  \end{tabular}
\end{table}

\subsection{Configurations for the Ablation Studies}
\label{app:ablation_training_configurations}

All ablation studies are conducted on JSC OpenML. Unless explicitly varied by an experiment, they use the shared training settings described at the beginning of this appendix. The four ablation studies are reported separately below.

\subsubsection{Single-Layer Width}

The single-layer-width ablation varies the number $N$ of trainable LUT6 nodes. The temperature is adjusted with model width according to Table~\ref{tab:app_single_layer_ablation}. These configurations are used to produce Figure~\ref{fig:openml_scale_ablation}.

\begin{table}[H]
  \caption{Configurations used for the single-layer-width ablation on JSC OpenML.}
  \label{tab:app_single_layer_ablation}
  \centering
  \small
  \begin{tabular}{@{}rcc@{}}
    \toprule
    Width $N$ & $\tau$ & Training steps \\
    \midrule
       50 & 3.6  & 200,000 \\
      100 & 4.5  & 200,000 \\
      250 & 7.9  & 200,000 \\
      540 & 16.0 & 200,000 \\
    1,000 & 21.3 & 200,000 \\
    2,000 & 30.0 & 200,000 \\
    4,000 & 57.0 & 200,000 \\
    \bottomrule
  \end{tabular}
\end{table}

\subsubsection{Network Depth}
The network-depth ablation fixes every LUT6 layer at a width of $N=2{,}000$ and varies the number of layers from one to four. The corresponding temperatures are listed in Table~\ref{tab:app_depth_ablation}. These configurations are used for the network-depth study in Section~\ref{sec:network_depth}.

\begin{table}[H]
\caption{Configurations used for the network-depth ablation on JSC OpenML.}
\label{tab:app_depth_ablation}
\centering
\small
\begin{tabular}{@{}rccc@{}}
\toprule
Number of layers & Width per layer & $\tau$ & Training steps \\
\midrule
1 & 2,000 & 30.0 & 200,000 \\
2 & 2,000 & 66.0 & 200,000 \\
3 & 2,000 & 93.0 & 200,000 \\
4 & 2,000 & 110.7 & 200,000 \\
\bottomrule
\end{tabular}
\end{table}

\subsubsection{GroupSum Training Temperature}

The temperature ablation fixes the DiffLUT-Net architecture to Ours (1000, 500) and varies only the GroupSum training temperature. The completed runs cover
\[
  \tau \in \{1,2,\ldots,80\},
\]
giving 80 completed runs. All other training settings remain fixed. The highest observed accuracy is obtained at $\tau=40$, as shown in Figure~\ref{fig:tau_accuracy_sweep}.

\begin{table}[H]
  \caption{Configuration used for the GroupSum temperature ablation on JSC OpenML.}
  \label{tab:app_tau_ablation}
  \centering
  \small
  \begin{tabular}{@{}lccc@{}}
    \toprule
    Configuration & Evaluated $\tau$ & Best $\tau$ & Training steps \\
    \midrule
    Ours (1000, 500) & $1$--$80$ & 40 & 200,000 \\
    \bottomrule
  \end{tabular}
\end{table}

\subsubsection{Second-Layer Width}

Table~\ref{tab:app_second_layer_ablation} lists the configurations used in Figure~\ref{fig:openml_ablation_three_corrected}. We fix the first-layer width at $N_1=2{,}000$ and vary the second-layer width $N_2$. The \texttt{None} configuration contains only the 2,000-node first layer, which connects directly to GroupSum.

\begin{table}[H]
  \caption{Configurations used for the second-layer-width ablation on JSC OpenML.}
  \label{tab:app_second_layer_ablation}
  \centering
  \small
  \begin{tabular}{@{}cccc@{}}
    \toprule
    First-layer width $N_1$ & Second-layer width $N_2$ & $\tau$ & Training steps \\
    \midrule
    2,000 & \texttt{None} & 30.0 & 200,000 \\
    2,000 & 2,000 & 66.0 & 200,000 \\
    2,000 & 1,000 & 39.0 & 200,000 \\
    2,000 & 500 & 23.7 & 200,000 \\
    2,000 & 250 & 13.4 & 200,000 \\
    \bottomrule
  \end{tabular}
\end{table}

\end{document}